\documentclass[iicol,pdflatex,sn-mathphys-num]{sn-jnl}

\usepackage{graphicx}%
\usepackage{multirow}%
\usepackage{amsmath,amssymb,amsfonts}%
\usepackage{amsthm}%
\usepackage{mathrsfs}%
\usepackage[title]{appendix}%
\usepackage{xcolor}%
\usepackage{textcomp}%
\usepackage{manyfoot}%
\usepackage{booktabs}%
\usepackage{algorithm}%
\usepackage{algorithmicx}%
\usepackage{algpseudocode}%
\usepackage{listings}%

\usepackage[normalem]{ulem} 
\usepackage[colorinlistoftodos]{todonotes}

\newif\ifrevision
\revisiontrue 

\theoremstyle{thmstyleone}%
\theoremstyle{thmstyletwo}%

\theoremstyle{thmstylethree}%

\begin{document}

\title[Hybrid Multi-Objective Evolutionary Algorithms for Service Placement]{Hybrid Multi-Objective Evolutionary Algorithms for Service Placement in the Computing Continuum: A Comparative Study with Genetic Traceability}


\author[1]{\fnm{Sergi} \sur{Vivo}}\email{sergi.vivo@uib.cat}

\author*[1]{\fnm{Carlos} \sur{Guerrero}}\email{carlos.guerrero@uib.es}

\author[1]{\fnm{Isaac} \sur{Lera}}\email{isaac.lera@uib.es}

\affil*[1]{\orgdiv{Computer Science Department}, \orgname{University of the Balearic Islands}, \orgaddress{\street{Crta. Valldemossa km 7.5}, \city{Palma}, \postcode{E07122}, \state{Illes Balears}, \country{Spain}}}


\abstract{This paper addresses multi-objective service placement in computing continuum environments through a collaborative hybrid island-model MOEA. The key innovation is not the design of a new general hybrid algorithm, but the systematic application and analysis of heterogeneous hybridization for this specific optimization domain through two independent experimental campaigns: a first one with four state-of-the-art MOEAs (NSGA-II, NSGA-III, U-NSGA-III, and SMS-EMOA), and a second one with a complementary hybrid configuration based on NSGA-II, MOEA/TS, and MOCPO, both co-evolving and periodically exchanging solutions. These designs enable complementary search behaviors across islands and are naturally aligned with the distributed edge-fog-cloud architecture of the computing continuum, facilitating scalable parallel execution. To evaluate the approach, we define two research hypotheses: (i) whether hybrid cooperation yields significant performance gains over standalone algorithms, and (ii) whether all constituent algorithms contribute equally to the final outcomes. We combine standard Pareto-front quality indicators (GD, IGD, HV, S, and STE) with a traceability-oriented analysis based on genetic load, which quantifies the contribution of each island to the evolved solutions. Across 30 independent runs, the hybrid method outperforms most of the standalone baselines, and statistical tests confirm significant improvements. Results also show non-uniform contributions among islands, providing interpretable evidence of effective hybrid cooperation.}




\maketitle

\section{Introduction}
\label{sec:introduction}

The increasing pervasiveness of cloud, fog, and edge resources has led to the emergence of a distributed and heterogeneous infrastructure often referred to as the \emph{computing continuum}. In this paradigm, services and data can be dynamically deployed across multiple tiers, from cloud data centers to edge devices, aiming to minimize latency, reduce energy consumption, and adapt to contextual constraints. One of the central challenges in this context is the \emph{service placement problem}, which consists in determining the optimal allocation of services to computational nodes, while satisfying multiple objectives such as performance, cost, and resource utilization. Solving this problem efficiently and at scale is crucial for enabling intelligent orchestration mechanisms in next-generation distributed systems.

In recent years, service placement has been addressed using deterministic heuristics, mixed-integer programming, and metaheuristic methods~\cite{brogi2020place}. Among them, evolutionary algorithms such as genetic algorithms (GAs) stand out for their flexibility, scalability, and effectiveness in multi-objective optimization~\cite{GUERRERO2022101094}. Their ability to explore large search spaces and approximate Pareto-optimal fronts makes them well suited to service placement in fog and edge environments.

Despite their popularity, most works in the field rely on a single evolutionary algorithm applied homogeneously across the entire population. Approaches that explore the potential of hybridization, where different algorithms or configurations cooperate to solve the same problem, remain underexplored in the specific context of service placement in the computing continuum. This gap in the literature motivates our study, which aims to investigate whether hybrid schemes can yield superior solutions when compared to traditional, monolithic GA-based strategies.

In any case, the main novelty of this manuscript is not the proposal of a new genetic operator or a new standalone MOEA, but a framework for collaborative hybrid island-model for service placement in computing continuum scenarios and two first solid applications and analysis with well-stablished algorithms. Therefore, our emphasis is placed on the motivation for hybridization in this domain, on the cooperative dynamics among heterogeneous state-of-the-art algorithms, and on the resulting optimization benefits under realistic deployment constraints.

In this work, we propose a hybrid multi-objective optimization framework based on the island paradigm. Our framework consists of deploying a set of distinct genetic algorithms, each evolving an independent subpopulation (island). Periodically, these islands exchange selected individuals with one another, promoting genetic diversity and enabling cooperation between different search strategies. 

This design choice is also motivated by the intrinsic distributed nature of the computing continuum. Since the hybrid model is organized as islands with periodic migration, each island can be executed on different nodes or tiers (edge/fog/cloud), enabling a natural parallel and scalable implementation that aligns with the target architecture itself.

To make the contributions explicit, this paper advances the state of the art through the following key innovations:

\begin{itemize}
    \item We provide a framework for collaborative hybrid island-model for the service placement in computing continuum environments.
    \item We perform a first systematic application and analysis of this collaborative hybrid island framework with periodic migration to exploit complementary search dynamics.
    \item We design two heterogeneous hybrid case studies that combine four state-of-the-art MOEAs (NSGA-II, NSGA-III, U-NSGA-III, and SMS-EMOA) in the first one, and both a classical high-quality baseline (NSGA-II) and recently proposed high-performing multi-objective optimizers (MOEA/TS, and MOCPO) in the second case study.
    \item We formulate the method in an architecture-aware manner, so that island-based optimization naturally maps onto distributed edge-fog-cloud infrastructures and supports scalable parallel execution.
    \item We establish a dual validation framework that integrates Pareto-front quality assessment and statistical significance testing with internal traceability analysis.
    \item We introduce a genetic load analysis that quantifies the temporal contribution of each island/algorithm and explains the cooperative dynamics of the hybrid model.
\end{itemize}

After defining these contributions, we formalize two research hypotheses to evaluate both global effectiveness and internal algorithmic contribution.

The first hypothesis aims to assess the effectiveness of the hybrid approach: Null Hypothesis ($H_0^1$), there is no statistically significant difference between the performance of the hybrid algorithm and that of the individual algorithms executed independently; Alternative Hypothesis ($H_1^1$), the hybrid algorithm achieves statistically significant improvements over the individual algorithms.

The second hypothesis seeks to investigate the relative contribution of each constituent algorithm within the hybrid model: Null Hypothesis ($H_0^2$), all individual algorithms contribute equally to the performance improvements observed in the hybrid algorithm; Alternative Hypothesis ($H_1^2$), the contribution of individual algorithms to the hybrid algorithm's performance is not uniform, i.e., some algorithms have a greater impact than others.




To validate these hypotheses, we conduct an experimental study involving two experimental campaigns using the previously defined case studies. We design two complementary analysis to evaluate our results. First, to assess the quality of the solutions produced by the hybrid algorithm, we compute five widely accepted performance metrics~\cite{panagant_comparative_2021}: Generational Distance (GD)~\cite{van1999multiobjective2}, Inverted Generational Distance (IGD)~\cite{10.1007/978-3-540-31880-4_35}, Hypervolume (HV)~\cite{10.1007/978-3-540-70928-2_64}, Spacing (S)~\cite{schott1995fault}, and Spacing-to-Extend (STE)~\cite{TEJANI2019425}. Each algorithmic configuration is executed multiple times under different random seeds to ensure statistical robustness, and the resulting data are analyzed using appropriate non-parametric statistical tests. 


In the second analysis block, we aim to understand the internal dynamics of the hybrid model. We calculate a metric, referred to as \emph{genetic load}, which quantifies the contribution of each island's population to the individuals currently residing in all islands. This metric allows us to trace the origin of high-quality solutions throughout the evolutionary process, measuring influence since the beginning of the optimization. This analysis offers deeper insights into the cooperative behavior and effectiveness of the hybridization mechanism, as well as the relative impact of each participating algorithm.


\section{Related work}

Hybrid genetic algorithms have emerged as a promising extension of traditional evolutionary techniques, as they enable the combination of diverse search operators, representations, or algorithmic paradigms to enhance exploration and exploitation capabilities. These hybrid approaches are particularly useful in complex multi-objective optimization problems, where leveraging the strengths of multiple strategies can lead to more robust and higher-quality solutions. 

There are several well-established paradigms for designing hybrid genetic algorithms, each offering distinct mechanisms for integrating complementary optimization strategies. They can be systematically classified according to two orthogonal dimensions~\cite{Talbi2002,6148272}: the level of coordination between components (low-level vs. high-level), and the execution flow (sequential vs. collaborative).

Low-level hybridization integrates external methods directly into the internal operations of the genetic algorithm, often modifying operators like mutation, crossover, or selection. A typical example is the \textit{memetic algorithm}, where local search is embedded after crossover to refine individuals. In contrast, high-level hybridization involves separate algorithms running either independently or with limited communication, such as in \textit{ensemble} or \textit{multi-solver} strategies.

In sequential hybrids (also known as \textit{relay hybrids}), different algorithms are executed one after another, with each stage refining the output of the previous one. This approach is often used when distinct optimization strategies are suitable for different stages of the search (e.g., global exploration followed by local exploitation). Collaborative hybrids, on the other hand, run multiple algorithms in parallel and allow some form of information exchange or cooperation. The \textit{island model} falls within this category, enabling different subpopulations to evolve simultaneously under different algorithms and share solutions periodically.

Considering these two orthogonal dimensions, the existing literature can be grouped into four main categories. Within the category of sequential and high-level approaches for resource management in the computing continuum, several studies have investigated hybridizations of different algorithms. Notable examples include combinations such as GA followed by PSO~\cite{9308549}, PSO integrated with NSGA-II~\cite{8611104}, the Hungarian algorithm combined with GA~\cite{8735711}, and k-means clustering used in conjunction with GA~\cite{8325027}, where the first and second methods are applied in sequence.

For high-level collaborative strategies, one of the most frequently adopted coordination mechanisms involves the exchange of candidate solutions during the optimization process. For instance, Ren et al.~\cite{10.1002/dac.4652} implemented an approach in which solutions are shared between a GA and an ACO algorithm. An alternative coordination approach consists of running several optimization techniques independently in parallel and later selecting the most promising outcomes from each. As an example, the works of Kabirzadeh et al.~\cite{10.23919/FRUCT.2017.8250177} and Rahbari et al.~\cite{8311595} explored the integration of four different metaheuristics (GA, PSO, ACO, and AS) using a test-and-select scheme to determine the final result.

Low-level collaborative techniques involve a deeper and more intricate integration between metaheuristics, often requiring a reengineering of the overall optimization workflow. In such strategies, the execution process is modified to interleave or embed operations from multiple heuristics. For instance, Li et al.~\cite{LiZSS20} enhanced the traditional GA structure by incorporating an intermediate procedure derived from the BAS evolution mechanism, aimed at refining individuals after selection, crossover, and mutation. Similarly, Yadav et al.~\cite{8929234} designed a GA-driven population evolution approach, augmented by embedding PSO-inspired particle movement to boost convergence speed. In contrast, Lin and Yang~\cite{8339513} implemented a generic optimization framework directed by the DMA, where genetic operations such as crossover and mutation are integrated into the DMA's execution logic. Additionally, Shamseddine et al.~\cite{9220179} introduced a machine learning component to compute solution fitness, thereby improving optimization effectiveness.

Lastly, the deployment of low-level cooperative strategies through relay-based execution remains uncommon in the literature, primarily due to the inherent complexity and limited applicability of such approaches~\cite{Talbi2002}. In fact, we did not identify any instances of this type of strategy applied to optimization within the computing continuum domain.

Moreover, to the best of our knowledge, prior work has not provided a systematic experimental analysis of collaborative island-based hybrid strategies specifically for service placement in computing continuum environments, which further motivates the focus and contribution of this paper.

Adopting a broader perspective, some works can also be considered hybrid approaches, as they coordinate the simultaneous optimization of different system components using multiple algorithms or heuristics. For instance, Benamer et al.~\cite{9363254} proposed the use of a grouping genetic algorithm (GGA) to determine service placement, while a particle swarm optimization (PSO) algorithm allocated resources within the fog nodes. A similar strategy was presented by Reddy et al.~\cite{REDDY2020102428}, where a GA was responsible for assigning resources to virtual machines, and a reinforcement learning mechanism predicted their duty cycles. In another example, Brogi et al.~\cite{8812204} employed the Monte Carlo method to model the uncertainty of a fog environment with fluctuating network parameters, applying a GA for optimization across the various generated scenarios.

Recent literature has also highlighted the relevance of hybrid genetic strategies in adjacent domains that share key characteristics with computing continuum optimization, namely heterogeneity, dynamic constraints, and competing objectives. In this context, Tursunboev et al.~\cite{tursunboev2025} proposed an evolutionary multi-objective framework (MOHFL) based on NSGA-II to manage heterogeneity in federated learning environments. By jointly optimizing model size and client selection, their approach balances communication costs and error rates, reinforcing the suitability of hybrid paradigms for dynamic workload partitioning across nodes with disparate computational and communication capabilities.

Furthermore, integrating genetic algorithms with domain-specific local search strategies or predictive models has shown significant promise in infrastructure-intensive environments. Research in hybrid energy systems~\cite{TURSUNBOEV2024122847} demonstrates how combining metaheuristics with advanced scheduling can solve large-scale dispatch and resource allocation problems.

The practical applicability of these hybrid genetic techniques is also reflected in critical scenarios such as autonomous smart cities and real-time industrial monitoring. By combining the global search capacity of genetic algorithms with the precision of local optimization, as evidenced by recent multi-objective scheduling studies~\cite{PALAKONDA2025126326}, near-optimal placements can adapt to real-time end-user mobility.

Despite the diversity of hybrid strategies explored in the literature, we did not find any previous work, in the context of computing continuum optimization, that applies a multi-heuristic island model where each island executes a different metaheuristic. While the island model is a well-established collaborative framework, existing studies typically adopt homogeneous populations using the same algorithm across islands~\cite{GUERRERO2024154}. 
In contrast, our study introduces a heterogeneous island-based hybrid genetic algorithm tailored for the computing continuum, where multiple heuristics cooperate by periodically exchanging solutions. This approach not only leverages the complementary strengths of different algorithms but also addresses the complex multi-dimensional optimization challenges inherent to this domain, thereby contributing a novel perspective to both hybrid metaheuristics and service placement in distributed architectures.

\section{Hybrid design}
\label{sec_hybriddesing}

Among the available hybridization schemes, we opted for an island model due to several compelling reasons. First, island-based evolutionary algorithms are inherently suitable for parallel and distributed computing environments, which aligns well with the decentralized nature of the fog and edge infrastructures targeted in our problem. This would allow straightforward implementations in the own infrastructure of the computing continuum~\cite{GUERRERO2024154}.

Second, they allow for the simultaneous exploration of diverse regions of the search space by maintaining isolated subpopulations governed by different evolutionary dynamics. This diversity often results in better global performance and a reduced risk of premature convergence. Third, the island model naturally supports the combination of heterogeneous genetic algorithms, enabling us to investigate the complementary strengths of different strategies within a unified framework. Moreover, the controlled migration of individuals across islands fosters a balance between exploration and exploitation, facilitating the sharing of promising solutions while preserving diversity. These characteristics make the island model a highly attractive and flexible approach for designing hybrid genetic algorithms in multi-objective, complex optimization scenarios such as service placement in the computing continuum.

From an experimental-motivation perspective, this heterogeneous island design is particularly suitable for computing continuum because the optimization environment itself is heterogeneous, distributed, and multi-scale. Cloud, fog, and edge tiers expose different resource profiles, latency conditions, and operational constraints; therefore, relying on a single search dynamic may bias the exploration of the placement space. By assigning different MOEAs to different islands and coordinating them through migration, the method explicitly mirrors this diversity and enables complementary optimization pressures to coexist. In addition, migration introduces a controlled mechanism to exchange high-quality solutions across islands, which is consistent with the need to transfer placement knowledge across continuum tiers while preserving local specialization.

To define the proposal of our hybrid genetic approach based on the island model, several design choices would made regarding the configuration and interaction of subpopulations. These decisions are governed by a set of parameters that control, among others, the number of islands, the population structure within each island, the migration mechanism, etc.

Let us denote by $n_{\text{islands}}$ the total number of islands or subpopulations in the model. In our proposal, each island runs a different multi-objective evolutionary algorithm (MOEA), such as for example NSGA-II, SMS-EMOA, etc. Each island $k \in {1, 2, \dots, n_{\text{islands}}}$ evolves independently and maintains a local population of size $p_{\text{k}}$. Therefore, the total population size of the hybrid algorithm is $p_{\text{total}} = \sum_{k=1}^{n_{\text{islands}}} p_{\text{k}}$.

Communication between islands occurs at regular intervals. Specifically, let $g_{\text{interval}}$ denote the number of generations between two consecutive migration events. Every $g_{\text{interval}}$ generations, each island selects and sends $m_{\text{out}}$ individuals (solutions) to one or more destination islands. The selection of the migrants are based on the selection policy, $\mathcal{S}^{(k)} : P^{(k)} \rightarrow \tilde{P}^{(k)}$, where $P^{(k)}$ is the current population of an island $k$ and $\tilde{P}^{(k)}$ is the subset of selected individuals. For example, a common strategy for migration is based on fitness, typically sending the best-ranked solutions in the local Pareto front. It is worth noting that the selection policy can be either specific to each island or shared across the entire algorithm.

To represent the migration topology among islands, we define a binary adjacency matrix $M \in {0,1}^{n_{\text{islands}} \times n_{\text{islands}}}$. An element $M_{i,j} = 1$ indicates that island $i$ sends solutions to island $j$, while $M_{i,j} = 0$ implies no direct communication. This matrix allows us to encode arbitrary migration patterns, such as ring, fully connected, or directional topologies.

Incoming individuals received by an island are integrated into its population using a replacement policy. The replacement policy is defined as $\mathcal{R}^{(k)}: P^{(k)} \bigcup^{k_i} \tilde{P}^{(k')} \rightarrow P'^{(k)}\; \forall k' \; \text{such that} \; M_{k,k'} = 1$, where $P^{(k)}$ is the current population of an island $k$, $\tilde{P}^{(k')}$ are the subsets of migrant individuals that receives island $k$ as specified the migration matrix, and $P'^{(k)}$ is the resulting set of solutions in island $k$ and size $p_k$. Accordingly to the selection policy, the replacement policy can be defined independently for each island or applied globally to all islands.  

Algorithm~\ref{alg:hybrid_island_model} summarizes the complete procedure of the proposed hybrid island-model MOEA using the notation introduced in this section.

\begin{algorithm}[t]
\caption{Collaborative hybrid island-model MOEA for service placement}
\label{alg:hybrid_island_model}
\begin{algorithmic}[1]
\Require $n_{\text{islands}}$, $\{\mathcal{A}^{(k)}\}$, $\{p_k\}$, $g_{\text{interval}}$, $m_{\text{out}}$, $M$, $\{\mathcal{S}^{(k)}\}$, $\{\mathcal{R}^{(k)}\}$, $G_{\max}$
\Ensure $P_{\text{final}} = \bigcup_{k=1}^{n_{\text{islands}}} P^{(k)}$
\For{$k \gets 1$ to $n_{\text{islands}}$}
    \State Initialize island population $P^{(k)}$ with size $p_k$
\EndFor
\For{$g \gets 1$ to $G_{\max}$}
    \For{$k \gets 1$ to $n_{\text{islands}}$ \textbf{in parallel}}
        \State $P^{(k)} \gets \mathcal{A}^{(k)}(P^{(k)})$
    \EndFor
    \If{$g \bmod g_{\text{interval}} = 0$}
        \For{$k \gets 1$ to $n_{\text{islands}}$}
            \State $\tilde{P}^{(k)} \gets \mathcal{S}^{(k)}(P^{(k)}, m_{\text{out}})$
        \EndFor
        \For{$j \gets 1$ to $n_{\text{islands}}$}
            \State $I^{(j)} \gets \emptyset$
            \For{$k \gets 1$ to $n_{\text{islands}}$}
                \If{$M_{k,j} = 1$}
                    \State $I^{(j)} \gets I^{(j)} \cup \tilde{P}^{(k)}$
                \EndIf
            \EndFor
            \State $P^{(j)} \gets \mathcal{R}^{(j)}(P^{(j)} \cup I^{(j)})$
            \State Keep exactly $p_j$ individuals in $P^{(j)}$
        \EndFor
    \EndIf
\EndFor
\State \Return $\bigcup_{k=1}^{n_{\text{islands}}} P^{(k)}$
\end{algorithmic}
\end{algorithm}

It is important to note that the parametrization of these configurable features can either remain static throughout the entire optimization process or dynamically evolve over time. Adaptive parameter control allows the algorithm to emphasize exploration during the early stages and gradually shift focus towards exploitation as the search progresses, thereby improving convergence and solution quality.

In summary, the formulation defined in our design allows for a flexible and modular hybridization strategy, where different algorithms can contribute their specific strengths, and solution sharing is governed by a configurable and interpretable interaction mechanism.

This design proposal is precisely one of the main contributions of this article, as it provides a general framework for constructing and analyzing heterogeneous hybrid optimization strategies in computing continuum scenarios. If the experimental results show that the hybrid framework can improve the performance of standalone algorithms, this opens a broad set of future research directions aimed at progressively designing more effective hybrid proposals. In particular, different combinations of algorithms, migration policies, island topologies, parameter-control mechanisms, and genetic traceability strategies could be explored to identify hybrid configurations with increasingly better convergence, diversity, and robustness properties.

\section{Problem formulation of the benchmarking optimization}
\label{sec:problem_formulation}

We address the problem of deploying services across a distributed infrastructure in the computing continuum. Given a set of applications $A$ to be deployed on a heterogeneous set of computational nodes $I$, the objective is to determine a valid assignment of applications to nodes that adheres to a set of operational constraints $C$ and simultaneously optimizes multiple performance metrics $O$.

The computing continuum integrates processing and storage resources located at different levels of the network hierarchy, ranging from centralized cloud data centers, through intermediate fog nodes, to edge devices located close to end users~\cite{9237349}. In our formulation, all such components are abstracted uniformly as computational nodes $i \in I$, regardless of their position in the hierarchy. Each node $i$ is described by its capacity in terms of computational resources, i.e.,  number of cores, denoted as $i_{\text{resources\_cores}}$, and memory resources, denote as $i_{\text{resources\_mem}}$. Likewise, each application $a \in A$ is characterized by its demand for those resources, represented by $a_{\text{consumption\_percentatge\_core}}$ and $a_{\text{consumption\_mem}}$ respectively.

Additionally,  devices are characterized by their power-related attributes. First, each node has a base power value ($i_{\text{base\_power}}$), which represents the minimum amount of power consumed when the node is powered on. This base power is always required whenever the node hosts at least one service.

Second, the power consumption of CPU and memory resources is modeled using specific multipliers ($i_{\text{cpu\_pw\_mult}}$ and $i_{\text{mem\_pw\_mult}}$), which scale the base power to estimate the maximum power consumption of each component. For example, if a node has a base power of $i_{\text{base\_power}} = 10W$, and its CPU can consume up to 75W, then its CPU power multiplier would be $i_{\text{cpu\_pw\_mult}} = 7.5$, since $75W = 7.5 \times 10W$.

Finally, since the CPU power consumption model usually adopts a piecewise function~\cite{7515731}, contrary to the memory model which adopts a linear function model, each node also includes a definition of this CPU power consumption function, $pw\_cpu_i(x)$, to represent this behavior.

\begin{gather}
pw\_cpu_i(x) = \\ =
\begin{cases}
a_1^{i_{pw\_cpu}} x + b_1^{i_{pw\_cpu}} & \text{if } x \in [x_0^{i_{pw\_cpu}}, x_1^{i_{pw\_cpu}}) \\
a_2^{i_{pw\_cpu}} x + b_2^{i_{pw\_cpu}} & \text{if } x \in [x_1^{i_{pw\_cpu}}, x_2^{i_{pw\_cpu}}) \\
\vdots \\
a_n^{i_{pw\_cpu}} x + b_n^{i_{pw\_cpu}} & \text{if } x \in [x_{n-1}^{i_{pw\_cpu}}, x_n^{i_{pw\_cpu}}]
\end{cases}
\label{eq_cpupower}
\end{gather}

The assignment of applications to infrastructure nodes follows a many-to-many model: an application can be deployed on multiple nodes, and a node can concurrently host multiple applications. To encode such mappings, we define a binary decision matrix $X$, where each entry $x_{a,i}$ corresponds to the deployment status of application $a \in A$ on node $i \in I$. If $x_{a,i} = 1$, the application is deployed on that node; otherwise, $x_{a,i} = 0$.

End users, denoted as the set $U$, access applications through their connections to the infrastructure (e.g., through gateways or edge devices). User requests are also many-to-many: each user may request one or more applications, and each application can be requested by multiple users. This interaction is captured by a binary matrix $R$ of dimensions $|A| \times |U|$, where $r_{a,u} = 1$ indicates that user $u \in U$ requests application $a \in A$, and $r_{a,u} = 0$ otherwise.

The optimization objectives are designed to reflect the priorities of infrastructure operators and system designers. These include metrics related to resource utilization, latency, and quality of service. 
In particular, we have considered three optimization objectives: minimizing the distance between users and services, balancing resource usage, and reducing power consumption.

First, the distance between users and a given application $a$, namely $distance(a)$ is calculated as the mean value of the network latency between each of these users and the closest application instance, i.e. the minimum latency between the user and all the application instances. We consider this objective to achieve low user perceived response times that is important in distributed systems with latency-sensitive applications. Our indicator is formally defined as:
\begin{gather}
distance(a)=\frac{\sum_{\forall u \in U} r_{a,u} \times \min_{\forall x_{a,i}=1} d(x_{a,i},u)}{\sum_{\forall u \in U} r_{a,u}}
\end{gather}
where $d(x_{i,a},u)$ is the network latency between the device in which the user $u$ is connected and the device/node $i$ in which the application $a$ is instantiated.
Finally, the average of all the application distances is considered as an indicator of the first optimization objective, $o_1 \in O$:
\begin{gather}
o_1 \Rightarrow \overline{distance} = \sum_{\forall a \in A} \frac{distance(a)}{|A|}
\end{gather}

Second, the balanced usage of the resources is calculated as the variance of the usage percentages across all the nodes. We consider this optimization to achieve load balancing across nodes, prevention of resource hotspots (overloaded nodes), and efficient resource utilization (avoiding underutilized nodes). This objective is particularly important in distributed systems where uneven load distribution can lead to performance bottlenecks and inefficient resource usage. For the sake of simplicity, we consider the resource limitation for allocating services into a device is the available memory. This assumption could be easily extended to other resources by considering several components instead of just one. Thus, we formally define the usage percentage of a given device $i$ as:
\begin{gather}
usage(i)=\frac{\sum_{\forall a \in A} x_{a,i} \times a_{\text{consumption\_mem}}}{i_{\text{resources\_mem}}}
\end{gather}
Finally, the variance of the usages for all nodes is the indicator to measure the second objective,  $o_2 \in O$:
\begin{gather}
o_2 \Rightarrow \mathrm{Var}(usage) = \frac{1}{|I|} \sum_{i \in I} (usage(i) - \overline{usage})^2 \\
\text{where } \overline{usage} = \frac{1}{|I|} \sum_{i \in I} usage(i)
\end{gather}

Third, the power consumption is calculated in terms of the memory and CPU usage of the devices. We consider this optimization because nodes are frequently embedded in edge locations with limited energy budgets. Additionally, this helps to achieve lower operational costs and carbon emissions. The power consumption of a given device $i$ is formally defined as:
\begin{gather}
power(i) = \\ =
\begin{cases}
0 & \text{if} \sum_{\forall a \in A} x_{a,i} = 0\\
base(i) + mem(i) + \\ + cpu(i) & \text{otherwise}
\end{cases}
\end{gather}
where the power consumption is 0 if the device does not allocate any service, or otherwise, the sum of the base power, the power of the CPU, and the power of the memory.

Regarding these three components, first, the base power consumed by a node is directly defined by its model features, i.e., the base power consumption of the nodes ($i_{\text{base\_power}}$):
\begin{gather}
base(i) = i_{\text{base\_power}}
\end{gather}

Second, the memory consumption model follows a linear model that considers that the power consumption generated by the memory is directly proportional to the usage percentage of the memory concerning the maximum power consumption of the memory:
\begin{gather}
mem(i) = i_{\text{base\_power}} \times i_{\text{mem\_power\_mult}} \times usage(i)
\end{gather}
where $i_{\text{base\_power}} \times i_{\text{mem\_power\_mult}}$ calculates the maximum power consumption of the memory and $usage(i)$ is the percentage of memory consumption in the node.

Third, the CPU consumption is modelled as a piecewise function model concerning the CPU usage of all of the cores in the node:
\begin{gather}
mem(i) = i_{\text{base\_power}} \times i_{\text{cpu\_power\_mult}} \times \\ \times  pw\_cpu_i(cpu\_usage(i))
\end{gather}
where
\begin{gather}
cpu\_usage(i) = \\ = \frac{\sum_{\forall a \in A} x_{a,i} \times a_{\text{consumption\_percentatge\_core}}}{i_{\text{resources\_core}}}
\end{gather}
The $i_{\text{base\_power}} \times i_{\text{cpu\_power\_mult}}$ calculates the maximum power consumption of the CPU when it has a 100\% usage, and the $pw\_cpu_i$ functions corresponds to the piecewise model of the CPU power consumption (Eq.\ref{eq_cpupower}) for the current CPU load, $cpu\_usage(i)$.

\section{Experimental Design}
\label{sec:experimental_design}

\subsection{Hybrid scheme parametrization}

To evaluate the effectiveness of the proposed hybrid genetic algorithm for service placement in the computing continuum, we conducted two independent experimental campaigns, both following the parametrization defined in Section~\ref{sec_hybriddesing}. In both campaigns, the hybrid scheme is specified by the same set of parameters: $n_{\text{islands}}$, number of islands; $p_{\text{island}}$, population size per island; $g_{\text{interval}}$, migration interval (in generations); $m_{\text{out}}$, number of solutions migrated from each island per migration round; $M$, migration matrix defining the communication topology; $\mathcal{S}^{(k)}$, the selection policy; and $\mathcal{R}^{(k)}$, the replacement policy.

The first campaign evaluates a four-island hybrid composed of established MOEAs (NSGA-II, NSGA-III, SMS-EMOA, U-NSGA-III), while the second campaign repeats the experimentation with an alternative hybrid configuration based on NSGA-II, MOEA/TS, and MOCPO. This second setting extends the empirical study with recently proposed high-performing optimizers, allowing us to assess whether the hybridization framework remains effective not only with classical MOEAs but also when incorporating newer-generation algorithmic components.

This experimental design is intended to test whether an optimization architecture that is itself heterogeneous and distributed can better match the characteristics of computing continuum than homogeneous single-algorithm configurations.

In the first experimental campaign, the algorithms, and therefore the number of islands, were chosen to provide a representative and diverse set of state-of-the-art MOEAs. Specifically, we considered four representative algorithms: NSGA-II (Algorithm~\ref{alg:island_nsgaii})~\cite{deb2002fast}, NSGA-III (Algorithm~\ref{alg:island_nsgaiii})~\cite{deb2014evolutionary}, SMS-EMOA (Algorithm~\ref{alg:island_smsemoa})~\cite{BEUME20071653}, and U-NSGA-III (Algorithm~\ref{alg:island_unsgaiii})~\cite{10.1007/978-3-319-15892-1_3}; hence, $n_{\text{islands}} = 4$. To improve reproducibility, Algorithms~\ref{alg:island_nsgaii}--\ref{alg:island_unsgaiii} summarize the island-level pseudocode adopted from their original formulations.

In the second experimental campaign, the algorithms, and therefore the number of islands, were selected to extend the evaluation with a mixed set of classical and recently proposed high-performing optimizers. Specifically, we considered NSGA-II (Algorithm~\ref{alg:island_nsgaii}) as a well-established baseline together with MOEA/TS (Algorithm~\ref{alg:island_moeats_project})~\cite{zhao2024moeats} and MOCPO (Algorithm~\ref{alg:island_mocpo_project})~\cite{adalja2025mocpo} as newer-generation algorithmic components; hence, $n_{\text{islands}} = 3$. This complementary configuration was introduced to determine whether the proposed hybridization framework remains effective when the cooperative search is driven not only by established MOEAs, but also by more recent optimization strategies.

Both experimental campaigns are analyzed independently throughout the paper. This separation is intentional, as each campaign addresses a different validation objective: the first one focuses on the behavior of the hybrid framework when combining established MOEAs, whereas the second one evaluates whether the same hybridization principles remain valid when incorporating newer-generation algorithms with a different solution interchange topology. In this way, the conclusions drawn from each campaign remain interpretable and are not confounded by mixing heterogeneous experimental settings into a single aggregate comparison.

NSGA-II serves as a foundational baseline, widely adopted for its effectiveness in maintaining diversity and convergence~\cite{deb2002fast}.
\begin{algorithm}[H]
\caption{Island-level NSGA-II procedure (adapted from~\cite{deb2002fast})}
\label{alg:island_nsgaii}
\begin{algorithmic}[1]
\Require Population $P_t$ of size $N$
\State Generate offspring $Q_t$ from $P_t$ using selection, crossover, and mutation
\State Build combined population $R_t \gets P_t \cup Q_t$
\State Compute non-dominated fronts $F_1, F_2, \ldots$ from $R_t$
\State Initialize $P_{t+1} \gets \emptyset$, $i \gets 1$
\While{$|P_{t+1}| + |F_i| \le N$}
    \State Assign crowding distance in $F_i$ and set $P_{t+1} \gets P_{t+1} \cup F_i$
    \State $i \gets i + 1$
\EndWhile
\State Sort $F_i$ by descending crowding distance and fill remaining slots of $P_{t+1}$
\State \Return $P_{t+1}$
\end{algorithmic}
\end{algorithm}

NSGA-III extends this family with a reference-point-based survival strategy for many-objective optimization~\cite{deb2014evolutionary}.
\begin{algorithm}[H]
\caption{Island-level NSGA-III procedure (adapted from~\cite{deb2014evolutionary})}
\label{alg:island_nsgaiii}
\begin{algorithmic}[1]
\Require Population $P_t$ of size $N$, predefined reference directions $Z^r$
\State Generate offspring $Q_t$ from $P_t$
\State Build combined population $R_t \gets P_t \cup Q_t$
\State Compute non-dominated fronts $F_1, F_2, \ldots$ from $R_t$
\State Fill $P_{t+1}$ with complete fronts until split front $F_\ell$
\State Normalize objective values and associate members of $F_\ell$ to reference directions
\While{$|P_{t+1}| < N$}
    \State Select an underrepresented reference direction
    \State Add one associated candidate (minimum perpendicular distance if niche is empty)
\EndWhile
\State \Return $P_{t+1}$
\end{algorithmic}
\end{algorithm}

SMS-EMOA introduces a steady-state selection criterion driven by hypervolume contribution~\cite{BEUME20071653}.
\begin{algorithm}[H]
\caption{Island-level SMS-EMOA procedure (adapted from~\cite{BEUME20071653})}
\label{alg:island_smsemoa}
\begin{algorithmic}[1]
\Require Population $P$ of size $N$
\While{termination criterion not met}
    \State Generate one offspring $y$ from selected parents in $P$
    \State Update set $R \gets P \cup \{y\}$
    \State Compute non-dominated sorting of $R$
    \State Identify worst front $F_w$
    \If{$|F_w| > 1$}
        \State Compute hypervolume contribution of each solution in $F_w$
        \State Remove solution with smallest contribution
    \Else
        \State Remove dominated solution according to front order
    \EndIf
    \State Set updated population as $P$
\EndWhile
\State \Return $P$
\end{algorithmic}
\end{algorithm}

U-NSGA-III keeps the NSGA-III survival principle while introducing tournament pressure during mating~\cite{10.1007/978-3-319-15892-1_3}.
\begin{algorithm}[H]
\caption{Island-level U-NSGA-III procedure (adapted from~\cite{10.1007/978-3-319-15892-1_3})}
\label{alg:island_unsgaiii}
\begin{algorithmic}[1]
\Require Population $P_t$ of size $N$, reference directions $Z^r$
\State Select parents by tournament-based mating selection
\State Generate offspring $Q_t$ by crossover and mutation
\State Build combined population $R_t \gets P_t \cup Q_t$
\State Apply NSGA-III non-dominated sorting and reference-direction-based survival
\State Use niching to complete $P_{t+1}$ up to size $N$
\State \Return $P_{t+1}$
\end{algorithmic}
\end{algorithm}

MOCPO is implemented in this project as an experimental multi-objective wrapper over the custom placement encoding, inspired by the Multi-Objective Crested Porcupines Optimization algorithm~\cite{adalja2025mocpo}. It combines elitist non-dominated survival with an information feedback matrix and four offspring generation strategies: visual, auditory, odor, and physical.
\begin{algorithm}[H]
\caption{Lightweight island-level MOCPO procedure implemented in this project (adapted from~\cite{adalja2025mocpo})}
\label{alg:island_mocpo_project}
\begin{algorithmic}[1]
\Require Population $P_t$ of size $N$
\State Select elite set $E_t$ from $P_t$ using non-dominated rank and crowding distance
\State Build feedback matrix $B$ from the average placement patterns of $E_t$
\State Initialize offspring set $Q_t \gets \emptyset$
\While{$|Q_t| < N$}
\State Select one strategy from \textsc{Visual}, \textsc{Auditory}, \textsc{Odor}, or \textsc{Physical}
\State Generate a placement matrix guided by $B$ and/or elite donors according to the selected strategy
\State Perturb the generated placement to preserve exploration and add it to $Q_t$
\EndWhile
\State Repair $Q_t$, remove duplicates, and evaluate offspring
\State Select $P_{t+1}$ from $P_t \cup Q_t$ using rank-and-crowding survival
\State \Return $P_{t+1}$
\end{algorithmic}
\end{algorithm}

MOEA/TS is implemented in this project as an experimental MOEA/TS-inspired wrapper over the custom placement encoding, following the three-state search idea proposed in~\cite{zhao2024moeats}. Its offspring generation alternates between convergence, diversity, and balance states selected from the current generation index.
\begin{algorithm}[H]
\caption{Lightweight island-level MOEA/TS procedure implemented in this project (adapted from~\cite{zhao2024moeats})}
\label{alg:island_moeats_project}
\begin{algorithmic}[1]
\Require Population $P_t$ of size $N$
\State Select search state $s_t \in \{\textsc{Convergence},\textsc{Diversity},\textsc{Balance}\}$ from the generation cycle
\If{$s_t = \textsc{Convergence}$}
\State Select elite set $E_t$ and extract frequent task--node placement patterns
\State Generate $Q_t$ by sampling placements biased toward those elite patterns
\ElsIf{$s_t = \textsc{Diversity}$}
\State Select diverse individuals from the ranked population
\State Generate $Q_t$ by favoring less-used nodes and exploratory mutations
\Else
\State Generate $Q_t$ by combining convergence offspring, diversity offspring, and standard mating offspring
\EndIf
\State Repair $Q_t$, remove duplicates, and evaluate offspring
\State Select $P_{t+1}$ from $P_t \cup Q_t$ using rank-and-crowding survival
\State \Return $P_{t+1}$
\end{algorithmic}
\end{algorithm}

In the first experimental campaign, all the algorithms use the same population size for both the stand-alone and the hybrid island executions, i.e, $p_k=400$. This results in a total population size of the hybrid alternative of $p_{total}=1600$; in the second campaign, the  population sizes of the stand-alone algorithms are equal to the total population size of the hybrid alternative with $p_{total}=300$. This results in island population sizes of $p_k=100$. The migration interval was established in $g_{intervals}=100$ generations in both campaigns. These parameters were established during a preliminary exploratory phase before the experimental optimization phase as part of the parameters calibration process~\cite{GIBBS2015226}.

Although a dedicated sensitivity analysis for hybrid island configurations is outside the scope of the objectives of this study, the selected parameter ranges are not arbitrary. They are consistent with our previous calibration studies on related service-placement scenarios using stand-alone~\cite{10.1145/3603166.3632547} and distributed genetic algorithms~\cite{GUERRERO2024154}, where population size, migration/exchange frequency, and communication settings were systematically varied and analyzed.

As a migration strategy, we adopt a fully connected exchange scheme, for the first experimental campaign, in which every island shares its entire population with all other islands during each migration period. This is modeled using the migration matrix $M$, where $M_{i,j} = 1$ for all $i \neq j$ and $M_{i,i} = 0$, meaning that each island $i$ sends individuals to every other island $j \in K$, with $i \neq j$. The second campaign addopts a ring-based communication scheme, where the population of NSGA-II is sent to MOCPO, the population of MOCPO is sent to MOEA/TS, and the population of MOEA/TS is sent to NSGA-II. Consequently the migration matrix $M$ is modeled as $M_{i,j} = 1$ for all $i$ except for $M_{NSGA2,MOCPO} = 1$, $M_{MOCPO,MOEATS} = 1$, and $M_{MOEATS,NSGA2} = 1$.

The number of individuals exchanged in each migration step is defined by the parameter $m_{\text{out}}$, which is set equal to the population size, i.e., $m_{\text{out}} = p_k$, meaning that the full population of each island is exchanged. The selection policy $\mathcal{S}^{(k)}$ is the same for all the islands and it is the identity function, i.e., all solutions are selected without filtering or ranking. 

Contrary, the replacement policies $\mathcal{R}^{(k)}$ differ across the islands. These replacement mechanisms are inherently tied to the way each algorithm ranks or selects solutions from best to worst. As a result, the selection pressure and criteria for integrating incoming migrants vary depending on the underlying strategy of the local optimization algorithm. In particular, the newly arrived individuals are merged with the existing population, and the $p_{\text{island}}$ best individuals are maintained, and the worst $(n_{islands} - 1) \times p_{\text{island}}$ individuals are removed,  according to their survivor selection mechanism: NSGA-II, where solutions are ranked by Pareto dominance, then sorted using crowding distance to preserve diversity~\cite{deb2002fast}; NSGA-III, where solutions are associated with predefined reference directions to guide selection toward a well-distributed Pareto front~\cite{deb2014evolutionary}; SMS-EMOA, where solutions are prioritized based on their contribution to the overall hypervolume, favoring those that expand the front~\cite{BEUME20071653}; U-NSGA-III, where solutions are associated with reference directions, and selection is based on proximity to underrepresented directions, aiming to improve distribution without using density-based measures~\cite{10.1007/978-3-319-15892-1_3}; MOCPO, where solutions are first ranked by non-dominated sorting and crowding distance, and an elite subset is used to build an information feedback matrix that biases the generation of new placements toward frequently selected task--node assignments while preserving diversity through strategy-specific perturbations~\cite{adalja2025mocpo}; MOEA/TS, where selection alternates among convergence, diversity, and balance states such as convergence extracts placement patterns from high-quality ranked solutions, diversity samples from underrepresented node-usage regions, and balance combines both sources with the configured mating operator before applying elitist rank-and-crowding survival~\cite{zhao2024moeats}.

\subsection{Infrastructure and scenario parametrization}

The two experimental campaigns were also defined over different scenario scales. In the first experimental campaign, the service-placement instances were generated with $|I|=50$ computational nodes, $|A|=50$ applications/tasks, and $|U|=25$ users. In the second experimental campaign, we considered a more compact configuration with $|I|=50$ computational nodes, $|A|=30$ applications/tasks, and $|U|=50$ users. These settings allow us to evaluate the proposed hybridization framework under two different problem sizes while preserving the same formal problem definition.

The objective of varying the two scenarios was not simply to scale the first scenario into a larger configuration, but rather to analyze the behavior of the algorithms under different workload distributions. For this reason, the second campaign increases the number of users while reducing the number of applications. As a result, each application receives a larger number of requests and must serve a higher number of users, creating a different service-demand pattern and a complementary evaluation setting for the proposed hybrid framework.

\subsection{Implementation}

The entire experimental framework was developed in Python, leveraging its flexibility and the availability of robust scientific libraries. The MOEAs used in this study were implemented using the pymoo library, which provides a modular and extensible environment for evolutionary computation~\cite{9078759}. Furthermore, the Wilcoxon test was implemented with the publicly available repository of the CEC 2013 test suite~\cite{Li2013}.

The developed framework is designed to be highly modular and parameterizable, allowing for the implementation and execution of any of the configurations defined in Section~\ref{sec_hybriddesing}. This flexibility refers to the number of islands, the communication between them, the definition of the policies, the optimization algorithms,\dots. This enables the exploration of alternative hybrid combinations with minimal effort. As such, the implementation itself constitutes an additional contribution of this work~\footnote{The source code and generated data in this work are available in a public repository at \url{https://github.com/acsicuib/GA-hierarchical-clustering}}.

\subsection{Performance Metrics}
\label{sec_performancemetrics}




We employ a set of well-established performance metrics for multi-objective optimization~\cite{panagant_comparative_2021} (GD, IGD, HV, S, and STE) to evaluate the first research hypothesis which investigates whether the hybrid algorithm outperforms the individual algorithms. These metrics collectively provide a comprehensive assessment of the quality of the solutions obtained. Specifically, GD and IGD measure the proximity of the obtained Pareto front to the true Pareto front, capturing convergence accuracy. HV quantifies both convergence and diversity by evaluating the dominated volume in the objective space. S and STE are diversity-oriented metrics that evaluate the uniformity and extent of the solution distribution across the Pareto front. The use of this diverse set of indicators ensures a balanced evaluation across convergence, diversity, and distribution, which is essential to rigorously assess whether the hybrid approach yields statistically significant improvements over the individual algorithms.

We introduce a custom metric called genetic load to validate the second research hypothesis, which states that the contribution of each algorithm to the final results is not equally balanced. This metric quantifies the influence that each evolutionary algorithm has had in the generation of a solution, by tracing the full ancestry of each individual in the final population. Each solution carries a vector of genetic load, which represents the proportion of its genetic material that originates from each of the algorithms involved in the hybrid process.

The genetic load is propagated during reproduction. When two parents generate an offspring, the child inherits a genetic load vector that is the arithmetic mean of the parents' vectors. Initially, all individuals in a population have a pure genetic load, that is, a 100\% contribution from their own algorithm. As generations progress and migration and recombination occur, these values evolve.

For example, if one parent has a pure genetic load from NSGA-II (codified with a genetic material vector such as [1, 0, 0, 0]) and the other is pure SMS-EMOA (a genetic load vector of [0, 1, 0, 0]), their offspring will have a genetic load vector of [0.5, 0.5, 0, 0]. In a more complex case, suppose one parent has a uniform genetic load across all four algorithms: [0.25, 0.25, 0.25, 0.25], and the other parent has [0.5, 0.3, 0.1, 0.1]. The resulting offspring will have a genetic load of [0.375, 0.275, 0.175, 0.175] meaning a 37.5\% of NSGA-II genetic load, 27,5\% of SMS-EMOA, 17.5\% of NSGA-III, and 17.5\% of U-NSGA-III.

This metric allows us to analyze the cumulative influence of each algorithm in the construction of the final solutions, helping us to identify dominant or underperforming strategies and to measure their relative impact throughout the evolutionary process.

We explicitly note that this formulation is an ancestry-based traceability metric (also referred to in our internal analysis as genetic workload), not a locus-level attribution metric. In line with prior ancestry-oriented analyses~\cite{11705:JCIS:2024:8}, our objective is to characterize historical algorithmic influence across the full evolutionary trajectory. Therefore, the metric quantifies how much each algorithm contributes to the lineage composition of final solutions, but it does not identify which specific decision variables were inherited from each parent, nor does it provide direct causal attribution at the gene level.

This distinction is important for interpretation: high genetic load/workload indicates persistent historical influence of an algorithm in the population dynamics, rather than guaranteed structural or causal contribution of particular genes. As future work, we plan to extend the traceability model with gene-level inheritance accounting and impact-weighted attribution, so that ancestry proportions can be complemented with variable-level contribution evidence.


\section{Results}

To assess the performance of the proposed hybridization framework, we analyze the results of the two experimental campaigns independently. In each campaign, the corresponding hybrid configuration is treated as the experimental group, while its constituent algorithms, executed individually and without any migration or hybridization mechanism, act as the control groups. Thus, in the first campaign, the hybrid model is compared against standalone executions of NSGA-II, NSGA-III, SMS-EMOA, and U-NSGA-III \footnote{Throughout the result figures of the first campaign, these configurations will be labeled as Hybrid, NSGA2, NSGA3, SMSEMOA, and UNSGA3, respectively.}; in the second campaign, the alternative hybrid model is compared against standalone executions of NSGA-II, MOEA/TS, and MOCPO\footnote{Throughout the result figures of the second campaign, these configurations will be labeled as Hybrid, NSGA2, MOEATS, and MOCPO, respectively.}. To preserve interpretability, the results of each campaign are discussed separately in different sections, avoiding direct aggregation of conclusions across heterogeneous experimental settings.

Due to the stochastic nature of evolutionary algorithms, each configuration was independently executed 30 times using different random seeds to ensure statistical robustness.

\subsection{First experimental campaign}

\subsubsection{First research hypothesis}

\begin{figure*}[t]
\centering
\includegraphics[width=0.9\textwidth]{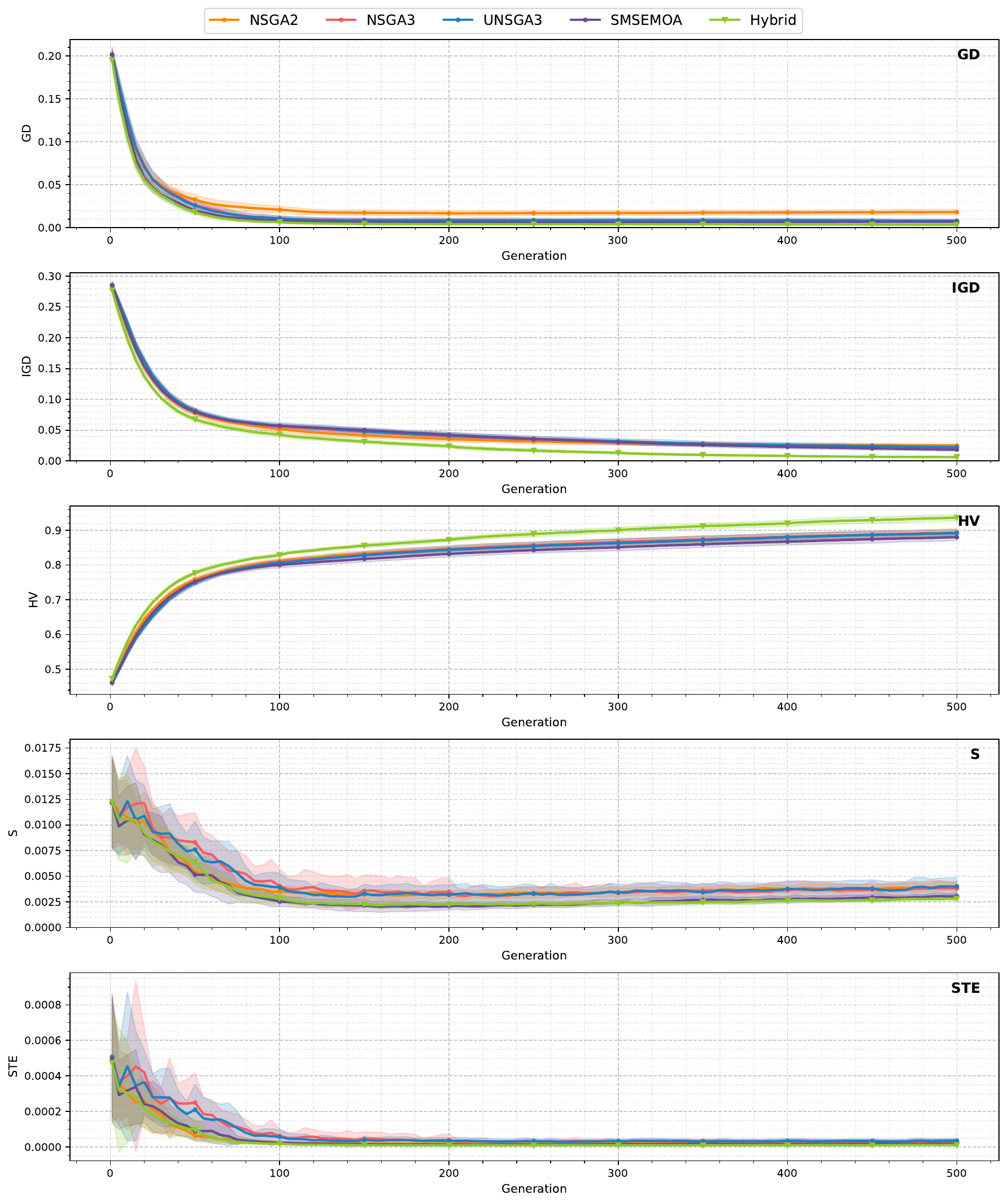}
\caption{First experimental campaign. Evolution of the values of the performance metrics GD, IGD, HV, S, and STE along the 500 generations for the four standalone optimizations (control groups) and the hybrid optimization (experimental group). The values of the metrics for each generation are calculated as the mean value of the 30 execution repetitions}\label{fig_metricsevolution}
\end{figure*}

We begin the analysis with a visual inspection of the results using two separate plots that illustrate the behavior of the five selected performance metrics across the different configurations. In Figure~\ref{fig_metricsevolution}, we show the evolution of each metric throughout the 500 generations of the optimization process. Since each configuration was executed 30 independent times, the plotted curves represent the average value of each metric across those 30 runs, providing a smoothed overview of the trends over time.

\begin{figure*}[h]
\centering
\includegraphics[width=0.98\textwidth]{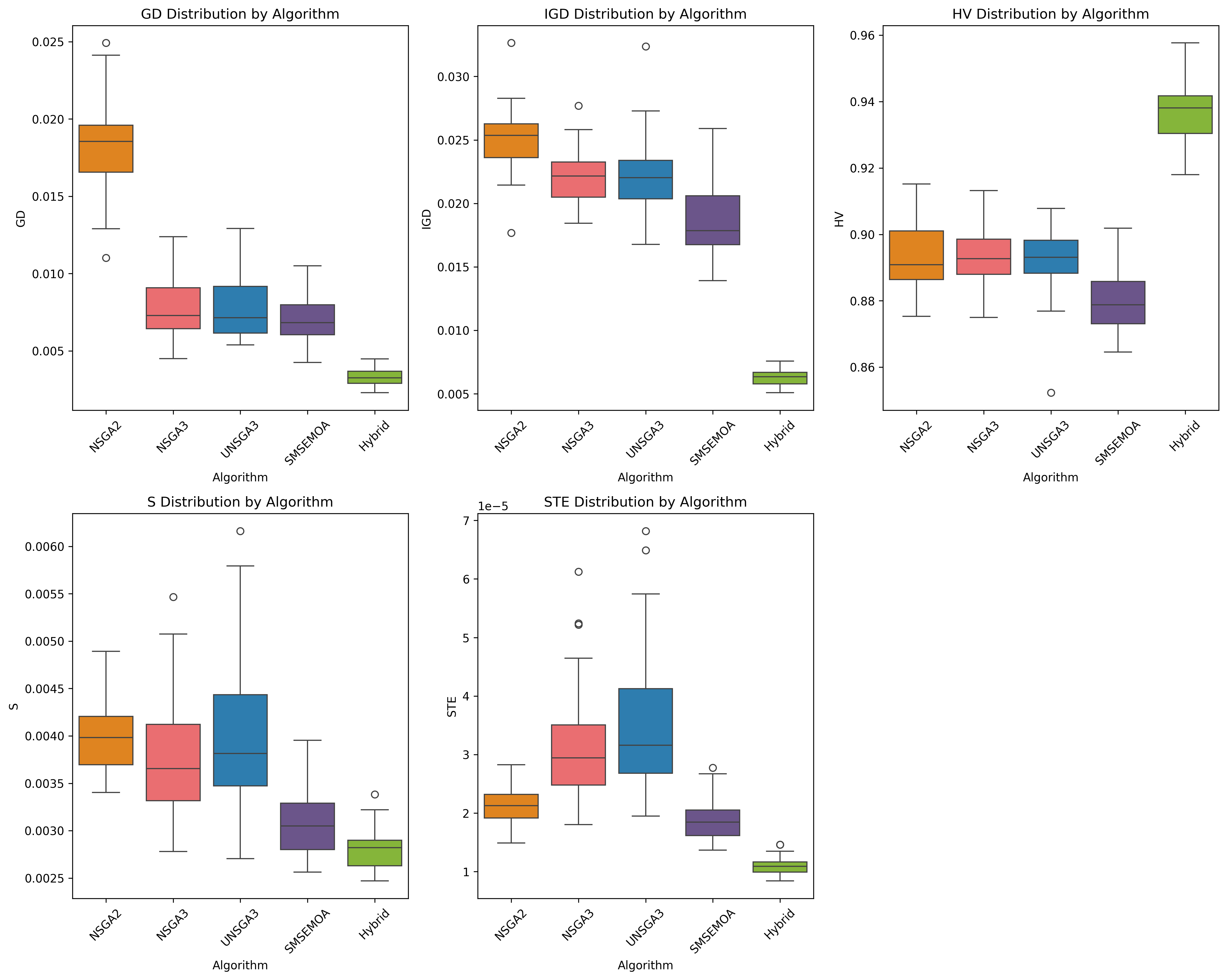}
\caption{First experimental campaign. Boxplots summarize the obtained results in generation 500 for the 30 execution repetitions for the four standalone optimizations (control groups) and the hybrid optimization (experimental group).}\label{fig_metricsboxplots}
\end{figure*}

Figure~\ref{fig_metricsboxplots} presents a more detailed view of the final performance by displaying boxplots for the values of the five metrics obtained in the last generation (generation 500). Each boxplot summarizes the distribution of metric values across the 30 independent runs for each of the four baseline algorithms and the hybrid configuration.

The first aspect worth highlighting is that, from a visual inspection of Figure~\ref{fig_metricsevolution}, all metrics appear to have stabilized by the final generation (500). Specifically, GD, S, and STE tend to converge earlier, around generation 200, whereas IGD and HV require more generations to reach a stable trend. This observation supports the choice of generation 500 as a representative point for comparing final results.

A second visual analysis of Figures~\ref{fig_metricsevolution} and~\ref{fig_metricsboxplots} reveals that the hybrid approach outperforms, in general terms, the individual algorithms across all five metrics. Specifically, for GD, IGD, S, and STE, lower values indicate better performance, while higher values are preferable for HV. In both figures, the hybrid scheme consistently achieves lower GD, IGD, S, and STE values, along with higher HV values, suggesting an overall improvement in solution quality, convergence, and distribution.


Beyond the visual analysis, a formal and objective statistical evaluation is necessary to assess the performance of the hybrid algorithm. Given that each algorithm, including the hybrid, was executed 30 times on the same set of optimization problems but with independent runs with different seeds, we adopted a paired non-parametric statistical test. Specifically, we employed the Wilcoxon Signed-Rank Test~\cite{wilcoxon1945individual} to compare the performance of the hybrid algorithm against each of the four baseline algorithms. This test is particularly suitable for paired experimental designs where the same problem instances are used across all algorithm executions, as it accounts for the relative performance differences without assuming normality of the data distribution. Statistical significance was assessed at a confidence level of $\alpha = 0.05$.



\begin{table}[h]
\footnotesize
\caption{Wilcoxon Signed-Rank Test results for the five performance metrics comparing the hybrid proposal with each control group (standalone algorithms).}\label{tab_wilcoxon}
\begin{tabular}{@{}lcc}
\toprule%
& \multicolumn{2}{@{}c@{}}{GD} \\\cmidrule{2-3}%
Hybrid vs. & $\rho$-value\footnotemark[1] & rank-biserial correlation ($r_{rb}$)\footnotemark[2] \\ 
\midrule
NSGA-II  & 0.000000002 & 1.000000000 \\ 
NSGA-III  & 0.000000002 & 1.000000000 \\ 
U-NSGA-III  & 0.000000002 & 1.000000000 \\ 
SMS-EMOA  & 0.000000002 & 1.000000000\\ 

\midrule

& \multicolumn{2}{@{}c@{}}{IGD} \\\cmidrule{2-3}%
Hybrid vs. & $\rho$-value\footnotemark[1] & rank-biserial correlation ($r_{rb}$)\footnotemark[2]  \\ 
\midrule
NSGA-II  & 0.000000002 & 1.000000000 \\ 
NSGA-III  & 0.000000002 & 1.000000000 \\ 
U-NSGA-III  & 0.000000002 & 1.000000000 \\ 
SMS-EMOA  & 0.000000002 & 1.000000000 \\ 

\midrule

& \multicolumn{2}{@{}c@{}}{HV} \\\cmidrule{2-3}%
Hybrid vs. & $\rho$-value\footnotemark[1] & rank-biserial correlation ($r_{rb}$)\footnotemark[2] \\ 
\midrule
NSGA-II  & 0.000000002 & 1.000000000 \\ 
NSGA-III  & 0.000000002 & 1.000000000 \\ 
U-NSGA-III  & 0.000000002 & 1.000000000 \\ 
SMS-EMOA  & 0.000000002 & 1.000000000 \\ 

\midrule

& \multicolumn{2}{@{}c@{}}{S} \\\cmidrule{2-3}%
Hybrid vs. & $\rho$-value\footnotemark[1] & rank-biserial correlation ($r_{rb}$)\footnotemark[2]  \\ 
\midrule
NSGA-II  & 0.000000002 & 1.000000000 \\ 
NSGA-III  & 0.000000009 & 0.993333333 \\ 
U-NSGA-III  & 0.000000002 & 1.000000000 \\ 
SMS-EMOA  & 0.001340603 & 0.817777778 \\ 

\midrule

& \multicolumn{2}{@{}c@{}}{STE} \\\cmidrule{2-3}%
Hybrid vs. & $\rho$-value\footnotemark[1] & rank-biserial correlation ($r_{rb}$)\footnotemark[2]  \\ 
\midrule
NSGA-II  & 0.000000002 & 1.000000000 \\ 
NSGA-III  & 0.000000002 & 1.000000000 \\ 
U-NSGA-III  & 0.000000002 & 1.000000000 \\ 
SMS-EMOA  & 0.000000002 & 1.000000000 \\ 

\botrule
\end{tabular}
\footnotetext[1]{Significant ($\rho < 0.05$): There is strong evidence that the two algorithms perform differently on the given metric. Not Significant ($\rho \ge 0.05$): There is insufficient evidence to conclude a difference in performance.}
\footnotetext[2]{Effect size (quantification of the magnitude of the difference): negligible ($|r_{rb}|<0.10$), small ($0.10 \leq |r_{rb}|<0.30$), medium ($0.30 \leq |r_{rb}|<0.50$), large ($|r_{rb}|\geq 0.50$). Max. value +1 (all ranks favor the hybrid algorithm). Min value -1 (all ranks favor the control algorithm).}
\end{table}

Table~\ref{tab_wilcoxon} reports the results of the Wilcoxon Signed-Rank Test for the five quality metrics under analysis. For each metric, both the p-value and the rank-biserial correlation (used as a measure of effect size) are presented. As shown in the table, the p-values are consistently and strongly favorable to the hybrid algorithm across all metrics, indicating statistically significant improvements compared to the individual baseline algorithms. Moreover, the rank-biserial correlation values are equal to 1 in most cases, which represents a perfect effect size, i.e., in all paired comparisons, the hybrid algorithm outperforms its counterparts without exception.

The only exception is observed for the S metric, where the rank-biserial correlation is slightly below 1.0 when comparing the hybrid approach with NSGA-III and SMS-EMOA.


The results of the Wilcoxon Signed-Rank Test yield a p-value very close to zero. This result does not indicate a truly null probability, but both a high consistency in the performance difference between the hybrid and baseline algorithms across all test instances (also contrasted with rank-biserial correlation values equal to 1) and an adequately large sample size (30 paired observations), which increases the statistical power of the test.

\begin{figure*}[h]
\centering
\includegraphics[width=0.98\textwidth]{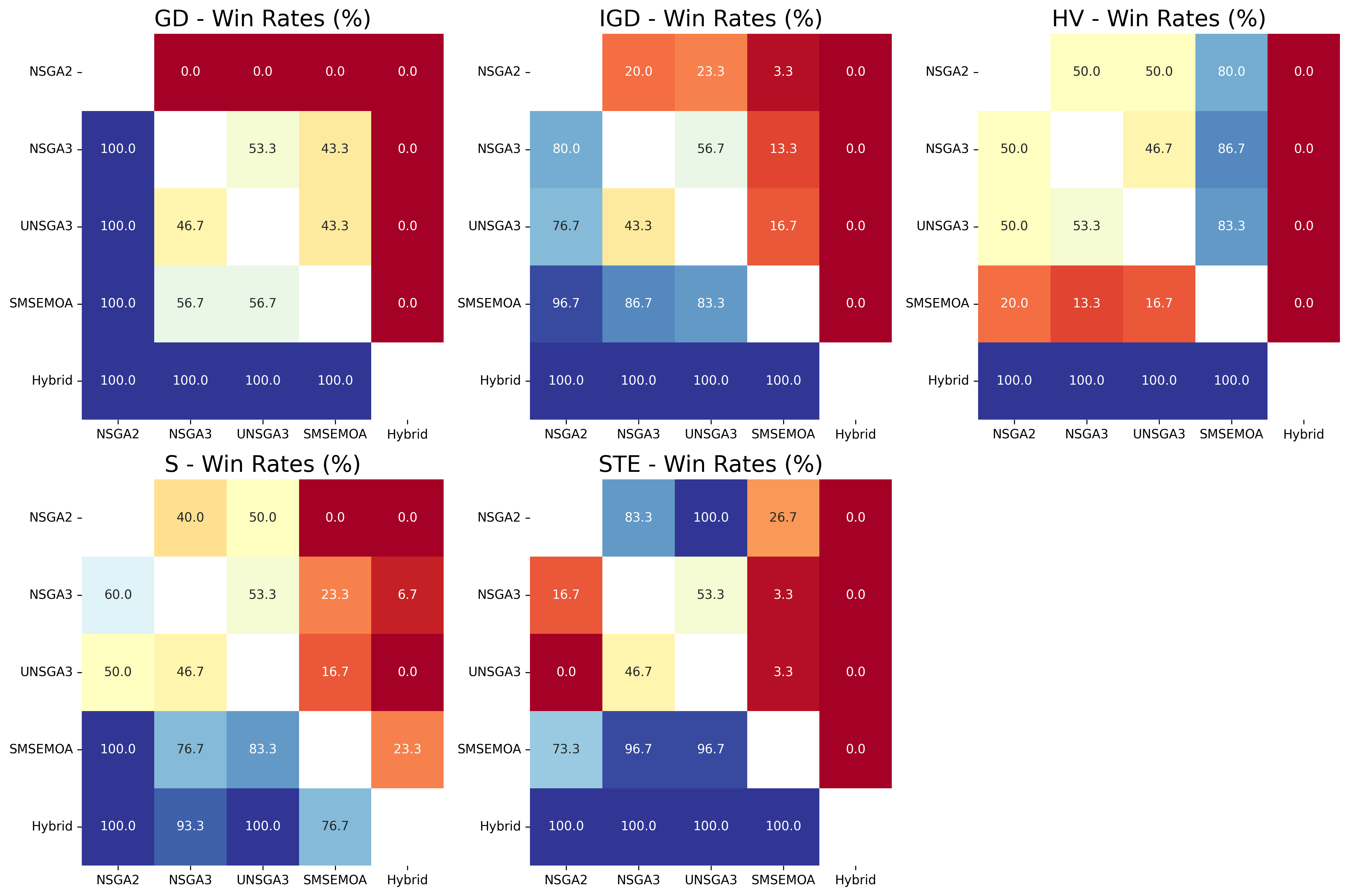}
\caption{First experimental campaign. Heatmap of win rates in pairwise comparisons: each cell shows the percentage of executions in which the row algorithm outperformed the column algorithm.}\label{fig_heatmap}
\end{figure*}

Figure~\ref{fig_heatmap} complements this analysis by providing a heatmap representation of the win rates observed in the pairwise comparisons. In this figure, each cell indicates the percentage of executions in which the algorithm on the corresponding row outperforms the algorithm on the corresponding column, according to the metric under analysis. These win rates offer a more intuitive and aggregated visualization of the dominance relationships between the hybrid algorithm and each of the baselines. As expected, the hybrid algorithm consistently shows high win rates against all individual algorithms, further reinforcing the statistical evidence of its superior performance.

Specifically, and in line with the results of the rank-biserial correlation, the only cases in which the hybrid algorithm does not outperform the baseline algorithms in 100\% of the pairwise comparisons occur for the S metric, in comparisons with NSGA-III and SMS-EMOA. In these two cases, NSGA-III achieves a better S value in 2 out of the 30 executions, and SMS-EMOA in 7 out of 30. Nevertheless, the hybrid algorithm still outperforms the four control algorithms in the remaining 70\% of the runs (21 out of 30), which confers it a better performance compare with any of the other algorithms.

For the sake of clarity, the Wilcoxon Signed-Rank Test analyses focused solely on comparing the hybrid algorithm with the control group algorithms, although all of them were computed to provide a broader perspective, and the complete set of results, including all pairwise win rate statistics, is available in the accompanying open-access repository containing all experimental data and source code. However, the win rate heatmap shown in Figure~\ref{fig_heatmap} also includes all pairwise comparisons between every algorithm to show some tips on the performance comparison between all the cases. Their inclusion in the heatmap allows for the visualization of lower win rates between algorithms other than the hybrid.


To sum up, the results demonstrate that the hybrid algorithm offers a statistically significant improvement over traditional algorithms across most metrics, both in terms of final solution quality and overall dominance. The use of non-parametric statistical tests ensures that these conclusions are robust to deviations from normality and outliers, providing a solid foundation for claims of algorithmic superiority. Therefore, these findings refuse the first null hypothesis ($H_0^1$) and support and validate the first alternative hypothesis ($H_1^1$) which states that the hybrid algorithm achieves statistically significant improvements over the individual algorithms.

\subsubsection{Second research hypothesis}

To assess the second hypothesis, we aim to determine whether any of the base algorithms used in the hybrid model exerts a greater influence on the resulting populations. To do so, we rely on the genetic load metric, previously introduced in Section~\ref{sec_performancemetrics}. This metric allows us to trace the ancestry of each solution within the hybrid population, quantifying the relative contribution of each island’s algorithm across generations. Specifically, the genetic load reflects not only which algorithm generated a particular solution, but also the algorithms responsible for generating its entire lineage. This enables a fine-grained analysis of the algorithmic influence patterns within the evolutionary dynamics of the hybrid model.

This evidence should be interpreted as historical influence over the evolutionary process (lineage-level contribution), and not as per-gene causal attribution.

\begin{figure*}[h]
\centering
\includegraphics[width=0.98\textwidth]{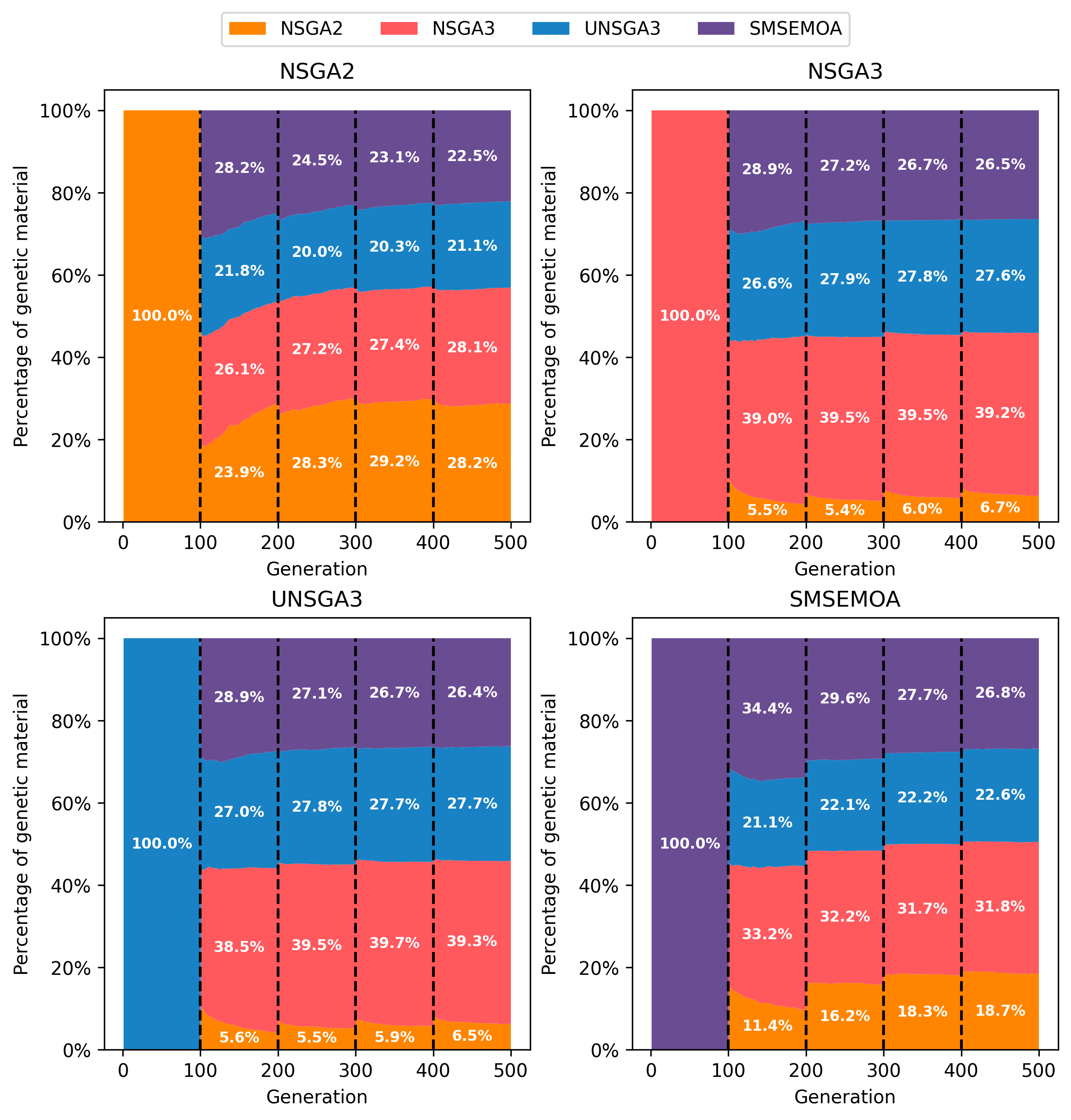}
\caption{First experimental campaign. Stacked area chart showing the evolution of the average genetic composition of the populations on each island throughout the optimization process. This visualization illustrates how the relative contribution of each algorithm to the genetic makeup of the populations changes over time. The percentage values correspond to the mean genetic load during each local optimization phase between two consecutive migration steps.}\label{fig_geneticload}
\end{figure*}

\begin{figure*}[h]
\centering
\includegraphics[width=0.98\textwidth]{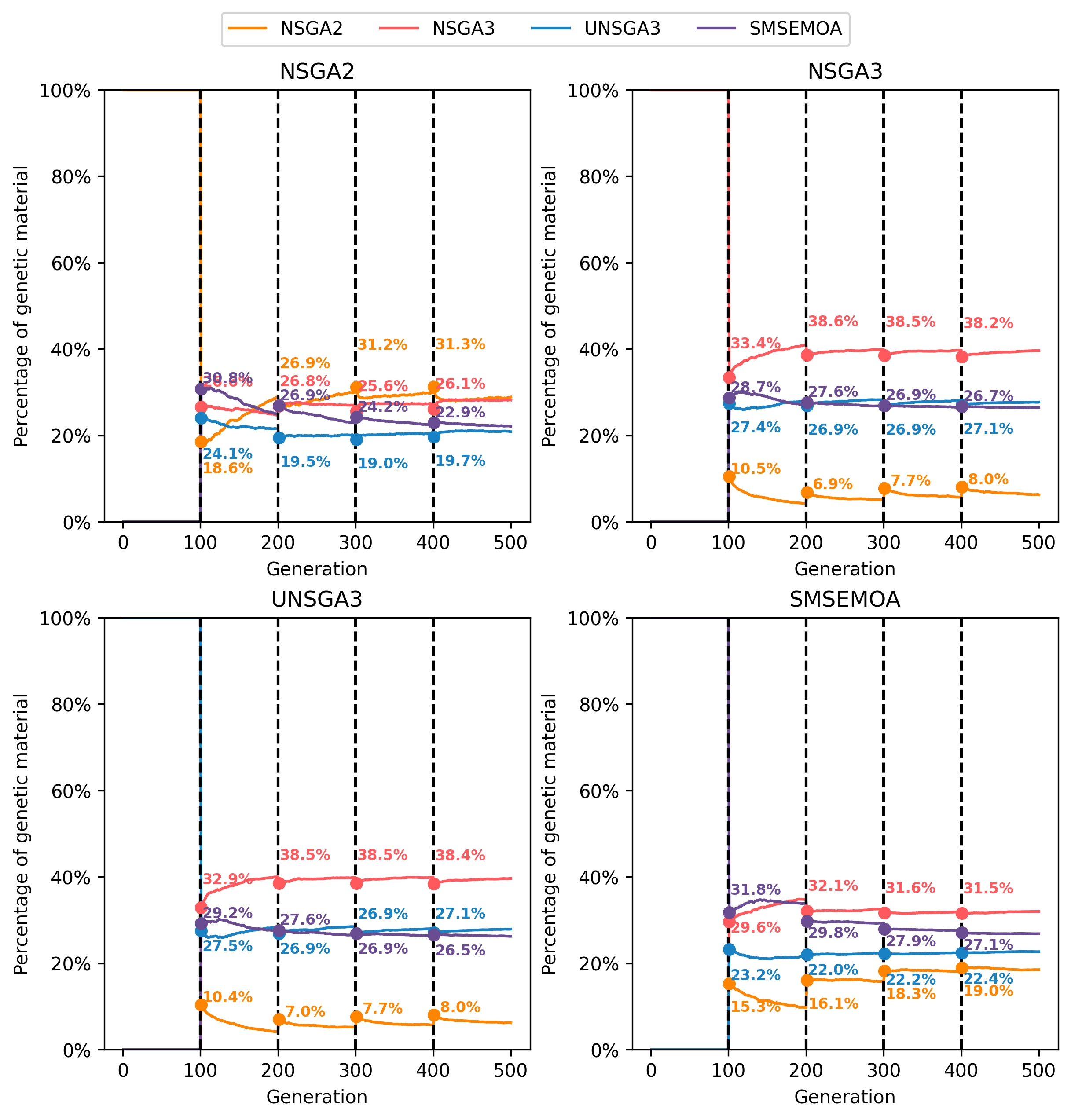}
\caption{First experimental campaign. Line chart showing the evolution of the average genetic composition of each population over time. Each line represents the average proportion of individuals originating from a specific algorithm, allowing a detailed view of how the genetic influence of each algorithm varies throughout the optimization process. The percentage values indicate the genetic load after each migration point, which are represented with vertical dashed lines.}\label{fig_geneticloadlines}
\end{figure*}

The evolution of the genetic load for each island across generations is displayed in Figure~\ref{fig_geneticload} and Figure~\ref{fig_geneticloadlines}. The first is a stacked area chart, which highlights the cumulative contribution of each algorithm's genetic material to the population over time, while the second is a multi-line chart that facilitates a more detailed comparison of the individual trends for each genetic load component. The total genetic load of an island's population is computed as the sum of the individual genetic loads of all the solutions within it. Additionally, vertical dashed lines indicate the migration points between islands, which occur every 100 generations. As expected, during the initial 100 generations, the populations of the four islands are genetically pure, meaning their solutions originate exclusively from the algorithm assigned to that specific island. After the first migration, however, the genetic composition of each population begins to change. Notably, abrupt shifts in genetic load can be observed at each migration point, while smoother, gradual transitions are visible between migrations. This reflects the combined effects of inter-island migration and the internal evolutionary dynamics within each island.

A general comparison of the genetic load across all islands reveals noticeable differences in the contribution and persistence of solutions generated by each algorithm. In particular, NSGA-III consistently exhibits the highest genetic load, indicating that its solutions are more likely to survive across generations and contribute offspring that remain in the population over time. This suggests a strong influence of NSGA-III in shaping the final Pareto front. In contrast, NSGA-II shows the lowest genetic load, meaning its individuals are less likely to persist or generate descendants that are retained in subsequent generations. This could reflect a lower adaptability or competitiveness of NSGA-II solutions in the heterogeneous environment of the island model, especially when compared to more recent or explorative algorithms.

When analyzing the genetic load from the perspective of each island, we observe significant differences in the diversity and balance of inherited solutions. The island running NSGA-II displays the most uniform genetic load, with contributions from all four algorithms ranging consistently between 18\% and 27\%. This suggests a balanced integration of solutions regardless of their origin, highlighting NSGA-II's tendency to maintain genetic diversity in a stable manner. A similar but slightly less uniform behavior is observed in the island using SMS-EMOA, where genetic contributions span from 15\% to 32\%, with more noticeable differences between the algorithms. In contrast, the islands running NSGA-III and U-NSGA-III exhibit the most uneven genetic distributions. In these populations, the influence of solutions originating from NSGA-II diminishes markedly, while NSGA-III's own solutions tend to dominate, suggesting a strong self-propagating effect. Interestingly, the genetic contribution from SMS-EMOA and U-NSGA-III remains relatively stable across all islands, indicating that their solutions maintain a consistent presence regardless of the algorithm employed locally.

In addition to the overall distribution of genetic load across islands, we are also interested in analyzing its temporal evolution at two distinct moments: (i) at the migration points (i.e., generations 100, 200, 300, etc.), and (ii) during the local optimization process between migrations, when each island runs its algorithm independently. By observing whether the genetic load of each algorithm increases or decreases during these intervals, we can better understand both the influence of the local optimization strategies and the impact of the migration process, including the selection and replacement policies applied to incoming individuals. This analysis allows us to distinguish which algorithms tend to propagate their solutions more effectively over time, and whether certain islands are more receptive or resistant to external genetic material.

In the first case, focusing on the abrupt changes in genetic load that occur at migration points, we observe a general trend: the genetic load associated with less dominant algorithms (such as NSGA-II) tends to increase, while the load from dominant sources (such as NSGA-III) tends to decrease. Although these variations are not very large, they suggest a homogenizing effect of the migration mechanism, which helps to counterbalance the dominance of certain genetic lineages and prevent premature convergence to suboptimal regions of the solution space. This contributes to maintaining diversity in the population, which is crucial for exploring a wider range of solutions. For the other two algorithms (U-NSGA-III and SMS-EMOA), the percentage of genetic load remains relatively stable across migration points, or exhibits only minor variations, indicating a more neutral or resilient behavior in the face of incoming genetic material.

The most significant changes in genetic load during the local optimization processes are observed in the second time period between migrations. This behavior indicates that early in the execution, the population is still highly dynamic, with rapid replacement and selection processes causing strong shifts in the dominance of certain genetic lineages. It also reflects that the initial migrations introduce new genetic material into relatively unstructured populations, making them more susceptible to influence. As the evolutionary process progresses, populations tend to stabilize, and the rate of change in genetic load becomes less pronounced, suggesting a gradual convergence and a reduced sensitivity to incoming migrants.

Similarly to the previous analysis on the migration points, we find consistent patterns across islands for the genetic loads evolution during the local optimization periods between migrations. This is true for all genetic loads except that of NSGA-II. The NSGA-II genetic load increases only in its native island, but decreases in the remaining three islands, suggesting that its solutions are not competitive enough to persist outside their original context. For the most prevalent genetic lineage, NSGA-III increases in all islands, reinforcing its dominance across the system. Additionally, U-NSGA-III genetic load tends to grow, while that of SMS-EMOA gradually declines. This suggests a general trend where NSGA-III and U-NSGA-III produce solutions better aligned with the selection and replacement dynamics of the system, whereas SMS-EMOA, despite early competitiveness, may generate solutions that lose influence over time.

In summary, the analysis of genetic load reveals clear patterns of dominance and persistence across the hybrid island model. From the perspective of genetic lineages, NSGA-III emerges as the most influential algorithm, contributing the largest share of long-lasting solutions in all populations, while NSGA-II exhibits the least persistence, with its genetic load declining in most islands except its own. U-NSGA-III shows moderate and stable influence, and SMS-EMOA, though initially competitive, tends to lose genetic presence over time. From the island-level viewpoint, the NSGA-II-controlled island stands out for maintaining the most balanced genetic diversity, while NSGA-III and U-NSGA-III islands show stronger internal dominance, particularly favoring their own solutions. Migration points slightly rebalance the genetic distribution by boosting underrepresented lineages, but local optimization phases reinforce the dominance of algorithms better adapted to the selection pressures of the system. 

These findings highlight the interplay between local algorithmic behavior and global cooperation mechanisms in shaping evolutionary dynamics within the hybrid model. These behaviors contributes to the generation of more diverse populations of solutions, which in turn enhances the exploration capabilities of the hybrid model. Moreover, the coexistence and interaction of distinct search strategies promote better convergence toward optimal regions, a more uniform distribution of solutions along the Pareto front, and ultimately result in higher-quality outcomes in terms of multi-objective performance.

These results allow us to address the second research hypothesis. The observed differences in genetic load dynamics clearly indicate that the contribution of each algorithm to the overall performance of the hybrid model is not uniform, thus supporting the alternative hypothesis ($H_1^2$). NSGA-III emerges as the most impactful algorithm, both in terms of the prevalence of its genetic material across islands and its ability to dominate during local optimization phases. Conversely, NSGA-II, despite maintaining balanced diversity in its own population, contributes less to the long-term evolution of solutions in the system as a whole. This asymmetric influence, consistently observed across multiple analytical perspectives, refutes the null hypothesis ($H_0^2$) and confirms that certain algorithms exert a significantly greater effect on the evolutionary dynamics and solution quality within the hybrid framework.





\subsection{Second experimental campaign}

\subsubsection{First research hypothesis}

\begin{figure*}[t]
\centering
\includegraphics[width=0.9\textwidth]{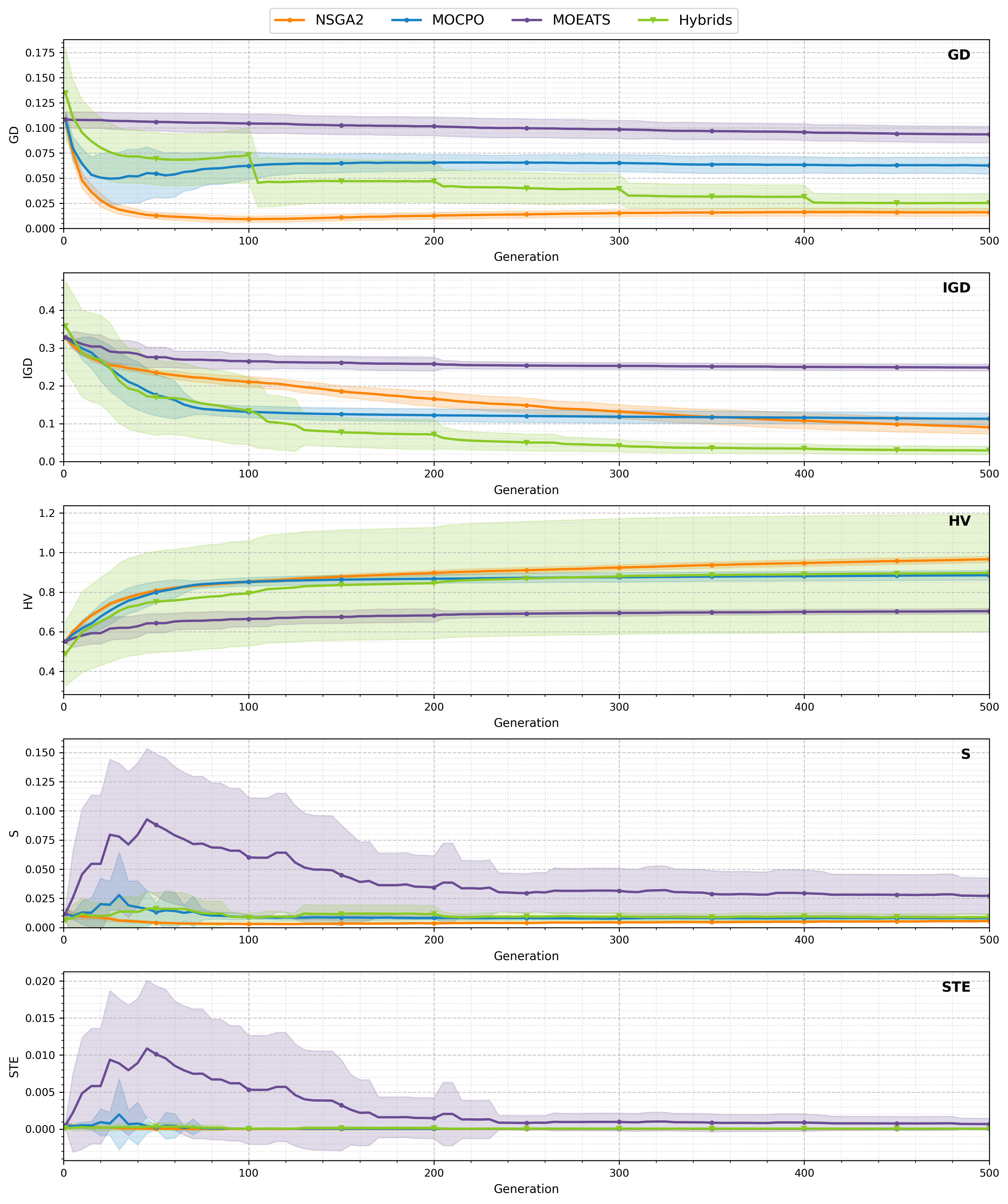}
\caption{Second experimental campaign. Evolution of the values of the performance metrics GD, IGD, HV, S, and STE along the 500 generations for the three standalone optimizations (control groups) and the hybrid optimization (experimental group). The values of the metrics for each generation are calculated as the mean value of the 30 execution repetitions}\label{fig_2metricsevolution}
\end{figure*}

\begin{figure*}[h]
\centering
\includegraphics[width=0.98\textwidth]{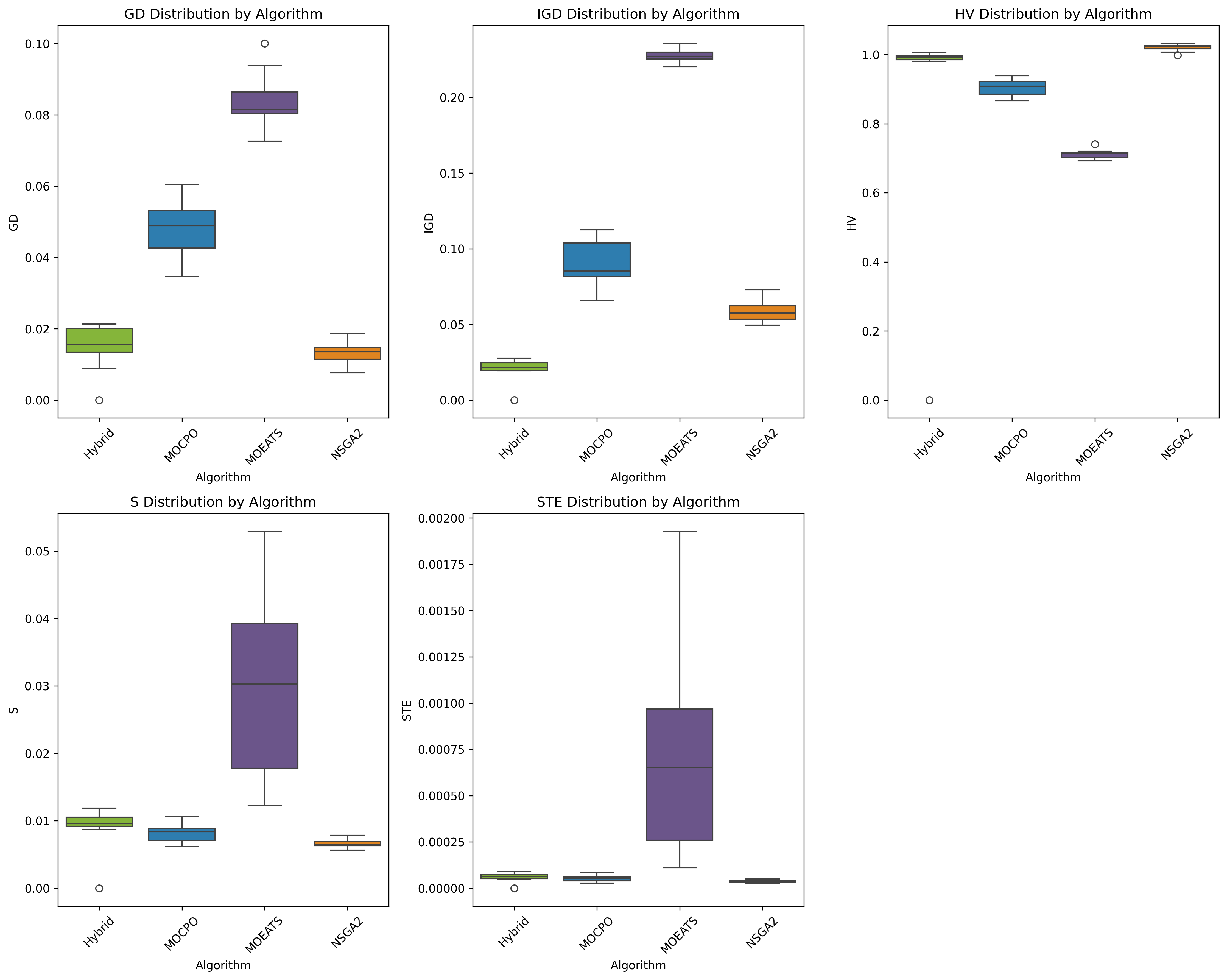}
\caption{Second experimental campaign. Boxplots summarize the obtained results in generation 500 for the 30 execution repetitions for the three standalone optimizations (control groups) and the hybrid optimization (experimental group).}\label{fig_2metricsboxplots}
\end{figure*}

The first aspect worth highlighting is that, from a visual inspection of Figure~\ref{fig_2metricsevolution}, most metrics appear to have stabilized by generation 300 in the second experimental campaign. Specifically, IGD, HV, S, and STE show relatively stable trends from that point onward, with only minor fluctuations during the remaining generations. The main exception is the GD of the hybrid configuration, which continues to decrease smoothly until generation 500, suggesting that this metric still benefits from additional search progress even after the rest of the indicators have largely converged. This observation supports the use of generation 500 as a representative point for comparing the final results of the second experimental campaign.

A second visual inspection of Figures~\ref{fig_2metricsevolution} and~\ref{fig_2metricsboxplots} shows that the hybrid approach either achieves the best performance or remains very close to the best standalone method across the evaluated metrics. In particular, for IGD and GD, the hybrid scheme emerges as the best-performing approach or stays very close to NSGA-II, which consistently ranks among the strongest algorithms for these indicators. A similar trend can be observed for HV, where the hybrid method also reaches values close to those obtained by NSGA-II. For the remaining metrics, S and STE, the hybrid algorithm shares the second position with only marginal differences with respect to MOCPO, further confirming its competitive and well-balanced behavior across convergence and distribution criteria.

\begin{table}[h]
\footnotesize
\caption{Second experimental campaign. Friedman Test results and mean ranks for the five performance metrics comparing the hybrid proposal with each control group (standalone algorithms).}
\label{tab_2friedman}
\centering
\begin{tabular}{@{}lllrr@{}}
\toprule
Metric & Algorithm & Mean  & Statistic\footnotemark[2] & $p$-value\footnotemark[3] \\
 &  & rank\footnotemark[1] &  &  \\
\midrule

\multirow{4}{*}{GD}
& NSGA2   & \textbf{1.4} & \multirow{4}{*}{27.12} & \multirow{4}{*}{0.000005556} \\
& MOCPO   & 3.0          &                        &                              \\
& MOEATS  & 4.0          &                        &                              \\
& Hybrids & 1.6          &                        &                              \\
\midrule

\multirow{4}{*}{IGD}
& NSGA2   & 2.0          & \multirow{4}{*}{30.00} & \multirow{4}{*}{0.000001380} \\
& MOCPO   & 3.0          &                        &                              \\
& MOEATS  & 4.0          &                        &                              \\
& Hybrids & \textbf{1.0} &                        &                              \\
\midrule

\multirow{4}{*}{HV}
& NSGA2   & \textbf{1.0} & \multirow{4}{*}{26.76} & \multirow{4}{*}{0.000006610} \\
& MOCPO   & 2.9          &                        &                              \\
& MOEATS  & 3.9          &                        &                              \\
& Hybrids & 2.2          &                        &                              \\
\midrule

\multirow{4}{*}{S}
& NSGA2   & \textbf{1.2} & \multirow{4}{*}{24.84} & \multirow{4}{*}{0.000016677} \\
& MOCPO   & 2.1          &                        &                              \\
& MOEATS  & 4.0          &                        &                              \\
& Hybrids & 2.7          &                        &                              \\
\midrule

\multirow{4}{*}{STE}
& NSGA2   & \textbf{1.4} & \multirow{4}{*}{21.36} & \multirow{4}{*}{0.000088622} \\
& MOCPO   & 2.2          &                        &                              \\
& MOEATS  & 4.0          &                        &                              \\
& Hybrids & 2.4          &                        &                              \\
\botrule
\end{tabular}
\footnotetext[1]{Mean Rank (relative performance ranking): A lower mean rank indicates a better performing algorithm (closer to 1st place across all datasets), while a higher mean rank indicates poorer performance. In this set of $4$ algorithms, values range from 1 to $4$.}
\footnotetext[2]{Friedman Statistic ($\chi^2_F$): Quantifies the overall variance between the mean ranks of the algorithms. A higher value indicates a greater discrepancy in performance across the group, leading to a lower $p$-value and higher statistical significance.}
\footnotetext[3]{Significant ($\rho < 0.05$): There is strong evidence that at least one algorithm performs differently from the others. Not Significant ($\rho \ge 0.05$): There is insufficient evidence to conclude any statistically significant difference in performance across the set of algorithms.}
\end{table}


We employed the Friedman Test~\cite{friedman1937use} (Table\ref{tab_2friedman}) to numerically determine whether there are statistically significant differences among the algorithms across the evaluated problem instances. The Friedman Test is particularly suitable for experimental settings involving more than two related methods, as it is based on the relative ranking of the algorithms within each block of observations and does not require the assumption of normality. Statistical significance was assessed at a confidence level of $\alpha = 0.05$.


Overall, the results obtained from the Friedman test are consistent with the trends previously identified through visual inspection. In particular, the statistical tests confirm the competitive advantage of the hybrid approach. Therefore, the numerical and graphical evidence jointly support the conclusions drawn from the visual analysis of the performance metrics.

To sum up, the results of the second experimental campaign indicate that, unlike the first experimental campaign, the hybrid algorithm exhibits more moderate advantages over the control experiments based on standalone algorithms. Although the hybrid approach still shows competitive behavior and maintains favorable performance across several metrics, the magnitude of its improvement is clearly smaller than that observed in the first campaign. This suggests that the benefits of hybridization, while still present, are more limited under the alternative algorithmic configuration considered in this second study. Therefore, these findings motivate a deeper comparative analysis, and in Section~\ref{sec_hybriddesigncomparison} we will investigate a possible explanation for this difference in results based on the distinct design characteristics of both hybrid algorithms campaign's designs.

\subsubsection{Second research hypothesis}

\begin{figure*}[h]
\centering
\includegraphics[width=0.98\textwidth]{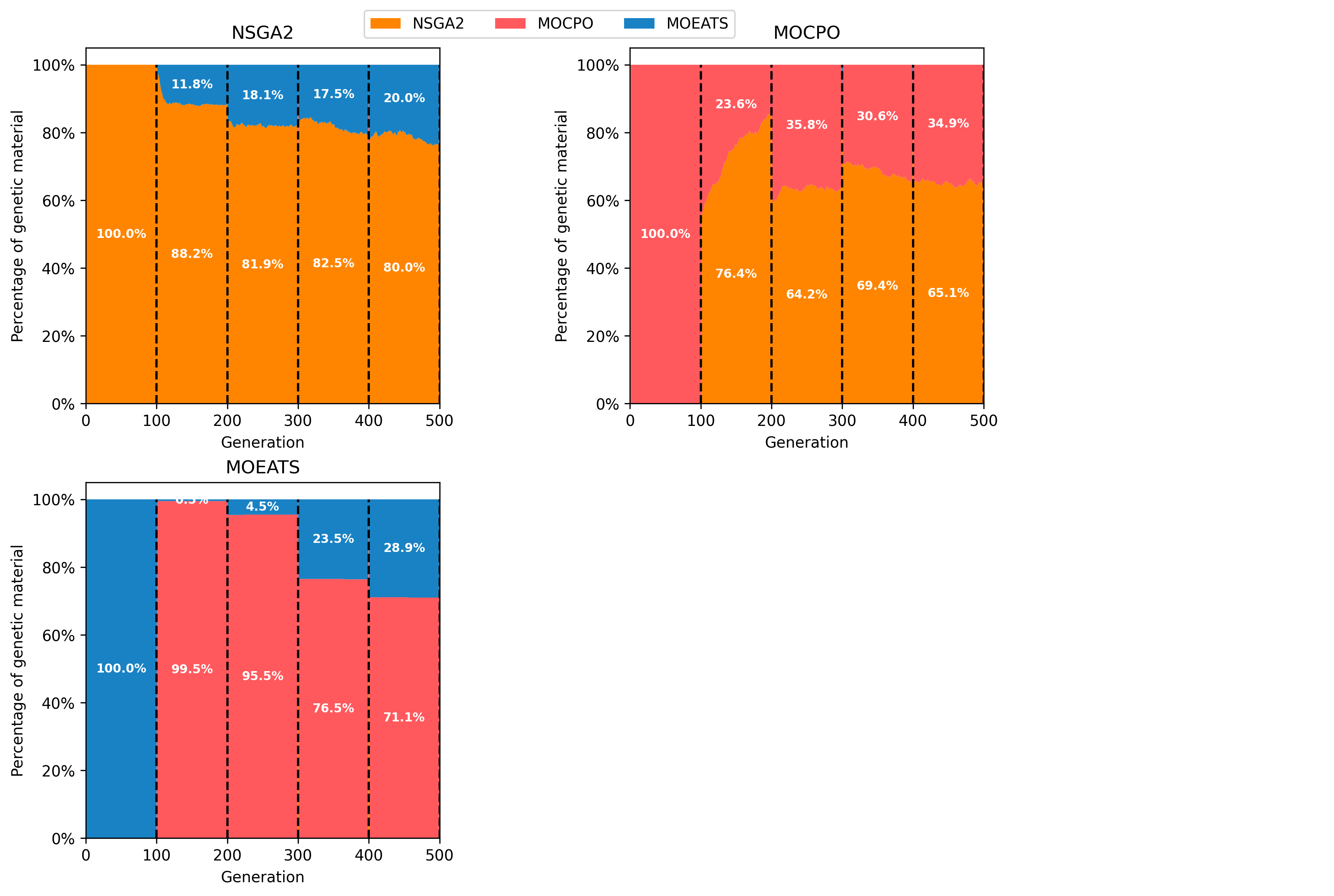}
\caption{Second experimental campaign. Stacked area chart showing the evolution of the average genetic composition of the populations on each island throughout the optimization process. This visualization illustrates how the relative contribution of each algorithm to the genetic makeup of the populations changes over time. The percentage values correspond to the mean genetic load during each local optimization phase between two consecutive migration steps.}\label{fig_2geneticload}
\end{figure*}


Contrary to the first experiment campaign, the second experimental campaign reveals a markedly different behavior for NSGA-II. In this case, NSGA-II-based solutions exhibit the highest prevalence not only within the NSGA-II island itself, but also within the MOCPO island, which is precisely the destination to which NSGA-II sends solutions at each migration step (Figure~\ref{fig_2geneticload}). This pattern suggests that, in the new hybrid configuration, NSGA-II generates individuals with a particularly high capacity to survive and remain competitive both in their island of origin and in the receiving island. Therefore, unlike the first campaign, where NSGA-II showed a limited genetic load, in the second hybrid design NSGA-II appears to play a much more influential role in sustaining the population dynamics and contributing persistent genetic material across islands.

It is worth noting, however, that this strong prevalence of NSGA-II tends to become progressively less pronounced as the generations evolve and as successive migration events take place between islands. Although NSGA-II-derived solutions dominate more clearly during the early stages of the evolutionary process, their relative predominance is gradually softened over time, suggesting that the continuous exchange of solutions promotes a more balanced redistribution of genetic influence across the hybrid system. This behavior indicates that, while NSGA-II plays a leading role in the initial consolidation of high-survival individuals, the ongoing interaction among islands allows the contribution of the other algorithms to increase as the search progresses.

Finally, the algorithm that appears to contribute the least in this second hybrid configuration is MOEA/TS. Its solutions show almost no prevalence within its own island during the initial stages of the search, and they are nearly entirely replaced from the very first generation. However, its prevalence gradually increases as new migration events take place and additional external solutions are incorporated. This behavior suggests that the initial solutions generated by MOEA/TS are comparatively weak, but also that the algorithm is able to take substantial advantage of incoming individuals produced by the other islands. In other words, once MOEA/TS is fed with externally generated solutions, it seems capable of refining them effectively through its own optimization strategy, which highlights a clear benefit of the hybrid design. Moreover, these progressively improved solutions also appear to be transferred back to the NSGA-II island over time, reinforcing the cooperative dynamics established between both components of the hybrid system.

Accordonly with results in the first experiment campaign, there are differences in genetic load dynamics which means different contribution of each algorithm to the overall performance of the hybrid model, thus supporting the alternative hypothesis ($H_1^2$), which confirms that certain algorithms exert a significantly greater effect on the evolutionary dynamics and solution quality within the hybrid framework. 

\subsection{Hybrid design comparison}
\label{sec_hybriddesigncomparison}

The first experimental campaign showed a general superiority of the hybred design with regard to the stand-alone results. On contrary, in the second experimental campaign, this superiority was limited to GD and IGD metrics. When an algorithm outperforms others in IGD but fails in Hypervolume (HV), Spacing, and STE, it suggests a high-quality but geographically restricted approximation of the Pareto Front. 

A superior IGD indicates that, on average, the solutions are very close to the true Pareto Front. However, IGD is a mean distance metric. If the algorithm finds a dense cluster of solutions in a specific, high-probability region of the front, the average distance to the reference set remains low. The failure in Spacing and STE (Spacing-to-Extent) reveals that these solutions are not only distributed unevenly (clumping) but are also confined to a narrow range. Even if these solutions are perfectly converged, they represent a narrow-minded view of the problem. Finalyy, Hypervolume requires the front to be both converged and extended to the boundaries. A low HV despite a good IGD is the classic signature of premature convergence to a sub-region of the Pareto Front.

The shift from a 4-algorithm all-to-all topology (first campaign) to a 3-algorithm ring topology (second campaign) explains why the latter achieves limited dominance in the metrics.

In a ring topology with only three algorithms, the flow of information is linear and slow. With fewer algorithms, there is less algorithmic dissonance. Each optimization paradigm explores the space differently. Only three algorithms might lack the specific operator needed to push the front toward the extreme boundary objectives. Moreover, the ring topology is a conservative migration strategy. It favors exploitation because it allows local islands to maintain their characteristics for longer, but in the second campaign, this likely led to neighborhood stagnation—where the islands agreed on a good enough region of the front and stopped pushing for the edges.

A four algorithms with a fully connected exchange mechanism creates a High-Intensity Information Cross-Pollination environment. Adding a fourth algorithm doesn't just add 25\% more power; it introduces a new set of search heuristics. This fourth perspective likely provides the necessary push to explore objective trade-offs that the other three were ignoring. An all-to-all exchange means that every island is constantly aware of the best solutions found across the entire search space. This prevents any single island from getting stuck in a local niche. Moreover, a fully connected topology, if one algorithm finds a breakthrough at the extremes of the front, that information is immediately available to all others. This creates a stretching effect on the population, directly improving STE and Hypervolume.

To sum up, the reason the 3-island ring version only won in IGD is that its conservative, low-diversity structure was excellent at refining a small portion of the front. It focused all its energy on polishing the solutions it already had, resulting in great proximity (GD/IGD) but zero breadth. In contrast, the 4-island all-to-all version creates a dynamic tension between the algorithms. The aggressive exchange forces the population to spread out to avoid being dominated by the diverse solutions arriving from all directions. This structural aggression fixes the Spacing and STE issues because it prevents the clumping of solutions, and it maximizes Hypervolume by ensuring that no objective corner is left unexplored.

\section{Conclusion}

This work presented a hybrid multi-objective optimization approach based on the island model, where each island runs a different MOEA. The main contribution lies in the definition of a collaborative hybrid framework, in the context of service placement in the computing continuum, that promotes both diversity and convergence through a structured migration process. Two case studies were defined for the experimental phase, integrating diverse algorithms such as NSGA-II, NSGA-III, SMS-EMOA, U-NSGA-III, MOCPO, and MOEA/TS. The proposal assesment was based not only on the final performance of the hybrid algorithm but also on the internal dynamics that emerge from the interaction between algorithms.

The results demostrate different behaviour for different hybrid designs. In the first experimental campaign, the results demonstrate that the hybrid model outperforms its constituent algorithms when executed independently, both in terms of quality and distribution of the Pareto front. The analysis of genetic load shows that NSGA-III exerts the strongest influence in generating enduring and dominant solutions, whereas NSGA-II has a more modest contribution, although it provides valuable diversity and stability in some islands. Genetic load trends across islands and over time suggest that the heterogeneous algorithmic landscape contributes to a more robust and balanced exploration–exploitation trade-off.

In the second experimental campaign, the results also show that the hybrid model preserves a competitive advantage over its constituent algorithms executed independently, although this advantage is more moderate than in the first campaign. The analysis of genetic load reveals a different internal dynamic, with NSGA-II exerting the strongest influence in generating solutions that persist both in its own island and in the MOCPO island, while MOEA/TS shows a weaker initial contribution but progressively increases its relevance as new migrated solutions are incorporated. These genetic load patterns across islands and generations suggest that the behavior of the hybrid design is strongly shaped by the specific interaction scheme established among its component algorithms, leading to a cooperative process in which some algorithms contribute more effectively to the generation of robust initial solutions, whereas others benefit more clearly from the refinement of imported individuals.

Furthermore, the evolution of genetic composition reveals that migration events promote the exchange of genetic material while avoiding premature convergence. Nonetheless, the effectiveness of migrations strongly depends on the replacement policy and the internal dynamics of each MOEA, suggesting that fine-tuning these strategies could further enhance performance.

Future work will explore the design of adaptive or learning-based migration strategies, including more sophisticated selection and replacement mechanisms for incoming individuals. Additionally, the incorporation of other state-of-the-art MOEAs or metaheuristic paradigms (e.g., decomposition-based or indicator-based methods) could lead to richer hybridization schemes. Finally, studying the impact of different topologies and communication frequencies between islands would help generalize the applicability of the proposed hybrid model. These new designs will be validated with well-established frameworks for hybrid optimization, such as AdaBoost- and clustering-based ensemble strategies~\cite{Palakonda2021ENMOEA,PALAKONDA2023120278}.

\backmatter

\section*{Declarations}

\bmhead{Funding}
This work was supported by Grants PID2021-128071OB-I00 and PID2024-158637OB-I00, both funded by MICIU/AEI/10.13039/501100011033 and by “ERDF A way of making Europe”, “ERDF/EU”.

\bmhead{Competing interest}

The authors declare that they have no known competing financial interests or personal relationships that could have appeared to influence the work reported in this paper.


\bmhead{Data and code availability}
The data generated in this study and the source code for the experimental phase are available in a public repository at \url{https://github.com/acsicuib/GeneticHybridizationFramework}, under the MIT License.

\bmhead{Author contribution}

All authors contributed to the conceptualization, methodology, validation, investigation and resources. Software, data curation and visualization were performed by Sergi Viv\'{o} and Isaac Lera. Methodology, formal analysis, supervision, project administration and funding acquisition were performed by Carlos Guerrero and Isaac Lera. The first draft of the manuscript was written by Carlos Guerrero and all authors commented on previous versions of the manuscript. All authors read and approved the final manuscript.

\bibliography{sn-bibliography}

\end{document}